# Fusing Multiple Multiband Images

Reza Arablouei and Frank de Hoog

*Abstract*—We consider the problem of fusing an arbitrary number of multiband, i.e., panchromatic, multispectral, or hyperspectral, images belonging to the same scene. We use the well-known forward observation and linear mixture models with Gaussian perturbations to formulate the maximum-likelihood estimator of the endmember abundance matrix of the fused image. We calculate the Fisher information matrix for this estimator and examine the conditions for the uniqueness of the estimator. We use a vector total-variation penalty term together with nonnegativity and sum-to-one constraints on the endmember abundances to regularize the derived maximum-likelihood estimation problem. The regularization facilitates exploiting the prior knowledge that natural images are mostly composed of piecewise smooth regions with limited abrupt changes, i.e., edges, as well as coping with potential ill-posedness of the fusion problem. We solve the resultant convex optimization problem using the alternating direction method of multipliers. We utilize the circular convolution theorem in conjunction with the fast Fourier transform to alleviate the computational complexity of the proposed algorithm. Experiments with multiband images constructed from real hyperspectral images reveal the superior performance of the proposed algorithm in comparison with the state-of-the-art algorithms, which need to be used in tandem to fuse more than two multiband images.

*Index Terms*—multiband image fusion; alternating direction method of multipliers; total variation; Cramer-Rao lower bound; maximum likelihood; linear mixture model.

## I. Introduction

THE WEALTH of spectroscopic information provided by hyperspectral images containing hundreds or even thousands of contiguous bands can immensely benefit many remote sensing and computer vision applications, such as object recognition [2], change detection [3], material classification [4], and spectral unmixing [5], commonly encountered in environmental monitoring, resource location, weather or natural disaster forecasting, etc. Therefore, finely-resolved hyperspectral images are in great demand [6]-[10]. However, limitations in light intensity as well as efficiency of the current sensors impose an inexorable trade-off between the spatial resolution, spectral sensitivity, and the signal-to-noise ratio (SNR) of existing spectral imagers [11]. As a results, typical spectral imaging systems can capture multiband images of high spatial resolution at a small number of spectral bands or multiband images of high spectral resolution with a reduced spatial resolution. For example, imaging devices onboard

Pleiades or IKONOS satellites[1] provide single-band panchromatic images with spatial resolutions of less than a meter and multispectral images with a few bands and spatial resolutions of a few meters while NASA's Airborne Visible/Infrared Imaging Spectrometer (AVIRIS)[2] provides hyperspectral images with more than two hundred bands but with a spatial resolution of several ten meters.

One way to surmount the abovementioned technological limitation of acquiring high-resolution hyperspectral images is to capture multiple multiband images of the same scene with practical spatial and spectral resolutions, then fuse them together in a synergistic manner. Fusing multiband images combines their complementary information obtained through multiple sensors that may have different spatial and spectral resolutions and cover different spectral ranges.

Initial multiband image fusion algorithms were developed to fuse a panchromatic image with a multispectral image and the associated inverse problem was dubbed pansharpening [12]-[15]. Most of the pansharpening algorithms are based on either of the two popular pansharpening strategies: component substitution (CS) and multiresolution analysis (MRA). The CS-based algorithms substitute a component of the multispectral image obtained through a suitable transformation by the panchromatic image. The MRA-based algorithm inject the spatial detail of the panchromatic image obtained by a multiscale decomposition into the multispectral image. There also exist hybrid methods that use both CS and MRA. Some of the algorithms originally proposed for pansharpening have been successfully extended to be used for fusing a panchromatic image with a hyperspectral image, a problem that is called hyperspectral pansharpening [15].

Recently, significant research effort has been expended to solve the problem of fusing a multispectral image with a hyperspectral one. This inverse problem is essentially different from the pansharpening and hyperspectral pansharpening problems since a multispectral image has multiple bands that are intricately related to the bands of its corresponding hyperspectral image. Unlike a panchromatic image that contains only one band of reflectance data usually covering parts of the visible and near-infrared spectral ranges, a multispectral image contains multiple bands each covering a smaller spectral range, some being in the shortwave-infrared (SWIR) region. Therefore, extending the pansharpening techniques so that they can be used to inject the spatial details of a multispectral image into a hyperspectral image is not straightforward. Nonetheless, an effort towards this end has led

R. Arablouei and F. de Hoog are with the Commonwealth Scientific and Industrial Research Organisation, Pullenvale QLD 4069 and Acton ACT 2601, Australia (email: reza.arablouei@csiro.au, frank.dehoog@csiro.au).

[1] http://www.satimagingcorp.com/satellite-sensors/
[2] http://aviris.jpl.nasa.gov/data/free_data.html

to the development of a framework called hypersharpening, which is based on adapting the MRA-based pansharpening methods to multispectral-hyperspectral image fusion. The main idea is to synthesize a high-spatial-resolution image for each band of the hyperspectral image by linearly combining the bands of the multispectral image using linear regression [16].

In some works on multispectral-hyperspectral image fusion, it is assumed that each pixel on the hyperspectral image, which has a lower spatial resolution than the target image, is the average of the pixels of the same area on the target image [17]-[21]. Clearly, the size of this area depends on the downsampling ratio. Based on this pixel-aggregation assumption, one can divide the problem of fusing two multiband images into subproblems dealing with smaller blocks and hence significantly decrease the complexity of the overall process. However, it is more realistic to allow the area on the target image corresponding to a pixel of the hyperspectral image to span as many pixels as determined by the point-spread function of the sensor, which induces spatial blurring. The downsampling ratio generally depends on the physical and optical characteristics of a sensor and is usually fixed. Therefore, spatial blurring and downsampling can be expressed as two separate linear operations. The spectral degradation of a panchromatic or multispectral image with respect to the target image can also be modeled as a linear transformation. Articulating the spatial and spectral degradations in terms of linear operations forms a realistic and convenient forward observation model to relate the observed multiband images to the target image.

Hyperspectral image data is generally known to have a low-rank structure and reside in a subspace that usually has a dimension much smaller than the number of the spectral bands [5], [22]-[24]. This is mainly due to correlations among the spectral bands and the fact that the spectrum of each pixel can often be represented as a linear combination of a relatively few spectral signatures. These signatures, called endmembers, may be the spectra of the material present at the scene. Consequently, a hyperspectral image can be linearly decomposed into its constituent endmembers and the fractional abundances of the endmembers for each pixel. This linear decomposition is called spectral unmixing and the corresponding data model is called the linear mixture model. Other linear decompositions that can be used to reduce the dimensionality of a hyperspectral image in the spectral domain are dictionary-learning-based sparse representation and principle-component analysis.

Many recent works on multiband image fusion, which mostly deal with fusing a multispectral image with a hyperspectral image of the same scene, employ the abovementioned forward observation model and a form of linear spectral decomposition. They mostly extract the endmembers or the spectral dictionary from the hyperspectral image. Some of the works use the extracted endmember or dictionary matrix to reconstruct the multispectral image via sparse regression and calculate the endmember abundances or the representation coefficients [25]. Others cast the multiband image fusion problem as reconstructing a high-spatial-resolution hyperspectral datacube from two datacubes degraded according to the mentioned forward observation model. When the number of spectral bands in the multispectral image is smaller than the number of endmembers or dictionary atoms, the linear inverse problem associated with the multispectral-hyperspectral fusion problem is ill-posed and needs to be regularized to have a meaningful solution. Any prior knowledge about the target image can be used for regularization. Natural images are known to mostly consist of smooth segments with few abrupt changes corresponding to the edges and object boundaries [26]-[28]. Therefore, penalizing the total-variation [29]-[31] and sparse (low-rank) representation in the spatial domain [32]-[35] are two popular approaches to regularizing the multiband image fusion problems. Some algorithms, developed within the framework of the Bayesian estimation, incorporate the prior knowledge or conjecture about the probability distribution of the target image into the fusion problem [36]-[38]. The work of [39] obviates the need for regularization by dividing the observed multiband images into small spatial patches for spectral unmixing and fusion under the assumption that the target image is locally low-rank.

When the endmembers or dictionary atoms are induced from an observed hyperspectral image, the problem of fusing the hyperspectral image with a multispectral image boils down to estimating the endmember abundances or representation coefficients of the target image, a problem that is often tractable (due to being a convex optimization problem) and has a manageable size and complexity. The estimate of the target image is then obtained by mixing the induced endmembers/dictionary and the estimated abundances/coefficients. It is also possible to jointly estimate the endmembers/dictionary and the abundances/coefficients from the available multiband data. This joint estimation problem is usually formulated as a non-convex optimization problem of nonnegative matrix factorization, which can be solved approximately using block coordinate-descent iterations [40]-[43].

To the best of our knowledge, all existing multiband image fusion algorithms are designed to fuse a *pair* of multiband images with complementary spatial and spectral resolutions. Therefore, fusing more than two multiband images using the existing algorithms can only be realized by performing a hierarchical procedure that combines multiple fusion processes possibly implemented via different algorithms as, for example, in [44] and [45]. In addition, there are potentially various ways to arrange the pairings and often it is not possible to know beforehand which way will provide the best overall fusion result. For instance, in order to fuse a panchromatic, a multispectral, and a hyperspectral image of a scene, one can first fuse the panchromatic and multispectral images, then fuse the resultant pansharpened multispectral image with the hyperspectral image. Another way would be to first fuse the multispectral and hyperspectral images, then pansharpen the resultant hyperspectral image with the panchromatic image. Apart from the said ambiguity of choice, such combined pair-wise fusions can be slow and inaccurate since they may require several runs of different algorithms and may suffer from propagation and accumulation of errors. Therefore, the increasing availability of multiband images with complementary characteristics captured by modern spectral imaging devices has brought about the demand for efficient and accurate fusion techniques that can handle multiple multiband images simultaneously.

In this paper, we propose an algorithm that can simultaneously fuse an arbitrary number of multiband images. We utilize the forward observation and linear mixture models to effectively model the data and reduce the dimensionality of the problem. Assuming matrix normal distribution for the observation noise, we derive the likelihood function as well as the Fisher information matrix (FIM) associated with the problem of recovering the endmember abundance matrix of the target image from the observations. We study the properties of the FIM and the conditions for existence of a unique maximum-likelihood estimate and the associated Cramer-Rao lower bound. We regularize the problem of maximum-likelihood estimation of the endmember abundances by adding a vector total-variation penalty term to the cost function and constraining the abundances to be nonnegative and add up to one for each pixel. The total-variation penalty serves two major purposes. First, it helps us cope with the likely ill-posedness of the maximum-likelihood estimation problem. Second, it allows us to take into account the spatial characteristics of natural images that is they mostly consist of piecewise plane regions with few sharp variations. Regularization with a vector total-variation penalty can effectively advocate this desired feature by promoting sparsity in the image gradient, i.e., local differences between adjacent pixels, while encourages the local differences to be spatially aligned across different bands [28]. The nonnegativity and sum-to-one constraints on the endmember abundances ensure that the abundances have practical values. They also implicitly promote sparsity in the estimated endmember abundances.

We solve the resultant constrained optimization problem using the alternating direction method of multipliers (ADMM) [46]–[51]. Simulation results indicate that the proposed algorithm outperforms several combinations of the state-of-the-art algorithms, which need be cascaded to carry out fusion of multiple (more than two) multiband images.

## II. DATA MODEL

### A. Forward observation model

Let us denote the target multiband image by $\mathbf{X} \in \mathbb{R}^{L \times N}$ where $L$ is the number of spectral bands and $N$ is the number of pixels in the image. We wish to recover $\mathbf{X}$ from $K$ observed multiband images $\mathbf{Y}_k \in \mathbb{R}^{L_k \times N_k}$, $k = 1, \dots, K$, that are spatially or spectrally downgraded and degraded versions of $\mathbf{X}$. We assume that these multiband images are geometrically co-registered and are related to $\mathbf{X}$ via the following forward observation model

$$\mathbf{Y}_k = \mathbf{R}_k \mathbf{X} \mathbf{B}_k \mathbf{S}_k + \mathbf{P}_k \tag{1}$$

where

$L_k \leq L$ and $N_k = N/D_k^2$ with $D_k$ being the spatial downsampling ratio of the $k$th image;

$\mathbf{R}_k \in \mathbb{R}^{L_k \times N}$ is the spectral response of the sensor producing $\mathbf{Y}_k$;

$\mathbf{B}_k \in \mathbb{R}^{N \times N}$ is a band-independent spatial blurring matrix that represents a two-dimensional convolution with a blur kernel corresponding to the point-spread function of the sensor producing $\mathbf{Y}_k$;

$\mathbf{S}_k \in \mathbb{R}^{N \times N_k}$ is a sparse matrix with $N_k$ ones and zeros elsewhere that implements a two-dimensional uniform downsampling of ratio $D_k$ on both spatial dimensions and satisfies $\mathbf{S}_k^\top \mathbf{S}_k = \mathbf{I}_N$;

$\mathbf{P}_k \in \mathbb{R}^{L_k \times N_k}$ is an additive perturbation representing the noise or error associated with the observation of $\mathbf{Y}_k$.

We assume that the perturbations $\mathbf{P}_k$, $k = 1, \dots, K$, are independent of each other and have matrix normal distributions expressed by

$$\mathbf{P}_k \sim \mathcal{MN}_{L_k \times N_k}(\mathbf{0}_{L_k \times N_k}, \mathbf{\Sigma}_k, \mathbf{I}_{N_k}) \tag{2}$$

where $\mathbf{0}_{L_k \times N_k}$ is the $L_k \times N_k$ zero matrix, $\mathbf{I}_{N_k}$ is the $N_k \times N_k$ identity matrix, and $\mathbf{\Sigma}_k \in \mathbb{R}^{L_k \times L_k}$ is a diagonal matrix that represents the correlation among rows of $\mathbf{P}_k$, which correspond to different spectral bands. Note that we consider the column-covariance matrices to be identity assuming that the perturbations are independent and identically-distributed in the spatial domain. However, by considering diagonal row-covariance matrices, we assume that the perturbations are independent in the spectral domain but may have non-identical variances at different bands.

Note that $\mathbf{Y}_k$, $k = 1, \dots, K$, in (1) contain the corrected (pre-processed) spectral values, not the raw measurements produced by the spectral imagers. The pre-processing usually involves several steps including radiometric calibration, geometric correction, and atmospheric compensation [52]. The radiometric calibration is generally performed to obtain radiance values at the sensor. The reflected sunlight passing through the atmosphere is partially absorbed and scattered through a complex interaction between the light and various parts of the atmosphere. The atmospheric compensation counters these effects and converts the radiance values into ground-leaving radiance or surface reflectance values. To obtain accurate reflectance values, one additionally has to account for the effects of the viewing geometry and sun's position as well as the surfaces structural and optical properties [6]. This pre-processing is particularly important when the multiband images to be fused are acquired via different instruments, from different viewpoints, or at different times. After the pre-processing, the images should also be co-registered.

### B. Linear mixture model

Under some mild assumptions, multiband images of natural scenes can be suitably described by a linear mixture model [5]. Specifically, the spectrum of each pixel can often be written as a linear mixture of a few archetypal spectral signatures known as endmembers. The number of endmembers, denoted by $M$, is usually much smaller than the spectral dimension of a hyperspectral image, i.e, $M \ll L$. Therefore, if we arrange $M$ endmembers corresponding to $\mathbf{X}$ as columns of the matrix $\mathbf{E} \in \mathbb{R}^{L \times M}$, we can factorize $\mathbf{X}$ as

$$\mathbf{X} = \mathbf{E} \mathbf{A} + \mathbf{P} \tag{3}$$

where $\mathbf{A} \in \mathbb{R}^{M \times N}$ is the matrix of endmember abundances and $\mathbf{P} \in \mathbb{R}^{L \times N}$ is a perturbation matrix that accounts for any possible inaccuracy or mismatch in the linear mixture mode. We assume that $\mathbf{P}$ is independent of $\mathbf{P}_k$, $k = 1, \dots, K$, and has a matrix normal distribution as

$$\mathbf{P} \sim \mathcal{MN}_{L \times N}(\mathbf{0}_{L \times N}, \mathbf{\Sigma}, \mathbf{I}_N) \tag{4}$$

where $\mathbf{\Sigma} \in \mathbb{R}^{L \times L}$ is its row-covariance matrix. Every column of $\mathbf{A}$ contains the fractional abundances of the endmembers at a pixel. The fractional abundances are nonnegative and often assumed to add up to one for each pixel.

The linear mixture model stated above has been widely used in various contexts and applications concerning multiband, particularly hyperspectral, images. Its popularity can mostly be attributed to its intuitiveness as well as relative simplicity and ease of implementation. However, there are a few caveats regarding this model that should be kept in mind. First, $\mathbf{X}$ in (3) corresponds to a matrix of corrected (pre-processed) values, not raw ones that would typically be captured by a spectral imager of the same spatial and spectral resolutions. However, whether these values are radiance or reflectance has no impact on the validity of the model, though it certainly matters for further processing of the data. Second, the model (3) does not necessarily require each endmember to be the spectral signature of only one (pure) material. An endmember may be composed of the spectral signatures of multiple materials or may be seen as the spectral signature of a composite material made of several constituent materials. Additionally, depending on the application, the endmembers may be purposely defined in particular subjective ways. Third, in practice, an endmember may have slightly different spectral manifestations at different parts of a scene due to variable illumination, environmental, atmospheric, or temporal conditions. This so-called endmember variability [53] along with possible nonlinearities in the actual underlying mixing process [54] may introduce inaccuracies or inconsistencies in the linear mixture model and consequently in the endmember extraction or spectral unmixing techniques that rely on this model. Moreover, the sum-to-one assumption on the abundances of each pixel may not always hold, especially, when the linear mixture model is not able to account for every material in a pixel possibly because of the effects of endmember variability or nonlinear mixing.

### C. Fusion model

Substituting (3) into (1) gives

$$\mathbf{Y}_k = \mathbf{R}_k \mathbf{E} \mathbf{A} \mathbf{B}_k \mathbf{S}_k + \breve{\mathbf{P}}_k \qquad (5)$$

where the aggregate perturbation of the $k$th image is

$$\breve{\mathbf{P}}_k = \mathbf{P}_k + \mathbf{R}_k \mathbf{P} \mathbf{B}_k \mathbf{S}_k.$$

Instead of estimating the target multiband image $\mathbf{X}$ directly, we consider estimating its abundance matrix $\mathbf{A}$ from the observations $\mathbf{Y}_k$, $k = 1, .. K$, given the endmember matrix $\mathbf{E}$. We can then obtain an estimate of the target image by multiplying the estimated abundance matrix by the endmember matrix. This way, we reduce the dimensionality of the fusion problem and consequently the associated computational burden. In addition, by estimating $\mathbf{A}$ first, we attain an unmixed fused image obviating the need to perform additional unmixing, if demanded by any application utilizing the fused image. However, this approach requires the prior knowledge of the endmember matrix $\mathbf{E}$. The columns of this matrix can be selected from a library of known spectral signatures, such as the U.S. Geological Survey digital spectral library[3], or extracted from the observed multiband images that have the appropriate spectral dimension.

## III. PROBLEM

### A. Maximum-likelihood estimation

In order to facilitate our analysis, we define the following vectorized variables

$$\mathbf{y}_k = \text{vec}\{\mathbf{Y}_k\} \in \mathbb{R}^{L_k N_k \times 1}$$
$$\mathbf{a} = \text{vec}\{\mathbf{A}\} \in \mathbb{R}^{MN \times 1}$$
$$\mathbf{p}_k = \text{vec}\{\breve{\mathbf{P}}_k\} \in \mathbb{R}^{L_k N_k \times 1}$$

where $\text{vec}\{\cdot\}$ is the vectorization operator that stacks the columns of its matrix argument on top of each other. Applying $\text{vec}\{\cdot\}$ to both sides of (5) while using the property $\text{vec}\{\mathbf{ABC}\} = (\mathbf{C}^{\top} \otimes \mathbf{A}) \text{vec}\{\mathbf{B}\}$ gives

$$\mathbf{y}_k = (\mathbf{S}_k^{\top} \mathbf{B}_k^{\top} \otimes \mathbf{R}_k \mathbf{E}) \mathbf{a} + \mathbf{p}_k \qquad (6)$$

where $\otimes$ denotes the Kronecker product.

Since $\mathbf{P}_k$ and $\mathbf{P}$ have independent matrix normal distributions [see (2) and (4)], $\mathbf{p}_k$ has a multivariate normal distribution expressed as

$$\mathbf{p}_k \sim \mathcal{N}_{L_k N_k}\left(\mathbf{0}_{L_k N_k}, \mathbf{I}_{N_k} \otimes \mathbf{\Sigma}_k + \mathbf{S}_k^{\top} \mathbf{B}_k^{\top} \mathbf{B}_k \mathbf{S}_k \otimes \mathbf{R}_k \mathbf{\Sigma} \mathbf{R}_k^{\top}\right)$$

where $\mathbf{0}_{L_k N_k}$ stands for the $L_k N_k \times 1$ vector of zeroes. Using the approximation $\mathbf{S}_k^{\top} \mathbf{B}_k^{\top} \mathbf{B}_k \mathbf{S}_k \approx \xi_k \mathbf{I}_{N_k}$ with $\xi_k > 0$, we get

$$\mathbf{p}_k \sim \mathcal{N}_{L_k N_k}\left(\mathbf{0}_{L_k N_k}, \mathbf{I}_{N_k} \otimes \mathbf{\Lambda}_k\right) \qquad (7)$$

where

$$\mathbf{\Lambda}_k = \mathbf{\Sigma}_k + \xi_k \mathbf{R}_k \mathbf{\Sigma} \mathbf{R}_k^{\top}.$$

In view of (6) and (7), we have

$$\mathbf{y}_k \sim \mathcal{N}_{L_k N_k}\left([\mathbf{S}_k^{\top} \mathbf{B}_k^{\top} \otimes \mathbf{R}_k \mathbf{E}] \mathbf{a}, \mathbf{I}_{N_k} \otimes \mathbf{\Lambda}_k\right).$$

Hence, the probability density function of $\mathbf{y}_k$ parametrized over the unknown $\mathbf{a}$ can be written as

$$f_{\mathbf{y}_k}(\mathbf{y}_k; \mathbf{a}) = \left|2\pi \mathbf{I}_{N_k} \otimes \mathbf{\Lambda}_k\right|^{-\frac{1}{2}}$$
$$\times \exp\left\{-\frac{1}{2}[\mathbf{y}_k - (\mathbf{S}_k^{\top} \mathbf{B}_k^{\top} \otimes \mathbf{R}_k \mathbf{E}) \mathbf{a}]^{\top}\right.$$
$$\left.(\mathbf{I}_{N_k} \otimes \mathbf{\Lambda}_k)^{-1} [\mathbf{y}_k - (\mathbf{S}_k^{\top} \mathbf{B}_k^{\top} \otimes \mathbf{R}_k \mathbf{E}) \mathbf{a}]\right\}.$$

Since the perturbations $\mathbf{p}_k$, $k = 1, \dots, K$, are independent of each other, the joint probability density function of the observations is written as

$$f_{\mathbf{y}_1, \dots, \mathbf{y}_K}(\mathbf{y}_1, \dots, \mathbf{y}_K; \mathbf{a}) = \prod_{k=1}^{K} f_{\mathbf{y}_k}(\mathbf{y}_k; \mathbf{a})$$
$$= \prod_{k=1}^{K} \left|2\pi \mathbf{I}_{N_k} \otimes \mathbf{\Lambda}_k\right|^{-\frac{1}{2}}$$
$$\times \exp\left\{-\frac{1}{2}\sum_{k=1}^{K} \left\|(\mathbf{I}_{N_k} \otimes \mathbf{\Lambda}_k^{-1/2})[\mathbf{y}_k - (\mathbf{S}_k^{\top} \mathbf{B}_k^{\top} \otimes \mathbf{R}_k \mathbf{E}) \mathbf{a}]\right\|^2\right\}$$

and the log-likelihood function of $\mathbf{a}$ given the observed data as

$$l(\mathbf{a}|\mathbf{y}_1, \dots, \mathbf{y}_K) = \ln f_{\mathbf{y}_1, \dots, \mathbf{y}_K}(\mathbf{y}_1, \dots, \mathbf{y}_K; \mathbf{a})$$
$$= -\frac{1}{2}\ln\left(\prod_{k=1}^{K}\left|2\pi \mathbf{I}_{N_k} \otimes \mathbf{\Lambda}_k\right|\right)$$
$$-\frac{1}{2}\sum_{k=1}^{K}\left\|(\mathbf{I}_{N_k} \otimes \mathbf{\Lambda}_k^{-1/2})[\mathbf{y}_k - (\mathbf{S}_k^{\top}\mathbf{B}_k^{\top} \otimes \mathbf{R}_k \mathbf{E})\mathbf{a}]\right\|^2.$$

---

[3] http://speclab.cr.usgs.gov/spectral.lib06/

Accordingly, the maximum-likelihood estimate of $\mathbf{a}$ is found by solving the following optimization problem

$$\hat{\mathbf{a}} = \arg\max_{\mathbf{a}} \, l(\mathbf{a}|\mathbf{y}_1, \dots, \mathbf{y}_K)$$
$$= \arg\min_{\mathbf{a}} \frac{1}{2}\sum_{k=1}^{K}\left\|(\mathbf{I}_{N_k} \otimes \mathbf{\Lambda}_k^{-1/2})[\mathbf{y}_k - (\mathbf{S}_k^{\top}\mathbf{B}_k^{\top} \otimes \mathbf{R}_k \mathbf{E})\mathbf{a}]\right\|^2. \tag{8}$$

This problem can be stated in terms of $\mathbf{A} = \text{vec}^{-1}\{\mathbf{a}\}$ as

$$\hat{\mathbf{A}} = \arg\min_{\mathbf{A}} \frac{1}{2}\sum_{k=1}^{K}\left\|\mathbf{\Lambda}_k^{-1/2}(\mathbf{Y}_k - \mathbf{R}_k \mathbf{E}\mathbf{A}\mathbf{B}_k\mathbf{S}_k)\right\|_{\text{F}}^2. \tag{9}$$

The Fisher information matrix (FIM) of the maximum-likelihood estimator $\hat{\mathbf{a}}$ in (8) is calculated as

$$\mathcal{F} = -\text{E}[\mathcal{H}_l(\mathbf{a})]$$

where $\mathcal{H}_l(\mathbf{a})$ denotes the Hessian, i.e., the Jacobian of the gradient, of the log-likelihood function $l(\mathbf{a}|\mathbf{y}_1, \dots, \mathbf{y}_K)$. The entry on the $i$th row and the $j$th column of $\mathcal{H}_l(\mathbf{a})$ is computed as

$$\frac{\partial^2}{\partial a_i \partial a_j} l(\mathbf{a}|\mathbf{y}_1, \dots, \mathbf{y}_K)$$

where $a_i$ and $a_j$ denote the $i$th and $j$th entries of $\mathbf{a}$, respectively. Accordingly, we can show that

$$\mathcal{F} = \sum_{k=1}^{K}(\mathbf{B}_k\mathbf{S}_k\mathbf{S}_k^{\top}\mathbf{B}_k^{\top} \otimes \mathbf{E}^{\top}\mathbf{R}_k^{\top}\mathbf{\Lambda}_k^{-1}\mathbf{R}_k\mathbf{E}).$$

If $\mathcal{F}$ is invertible, the optimization problem (8) has a unique solution given by

$$\hat{\mathbf{a}} = \left[\sum_{k=1}^{K}(\mathbf{B}_k\mathbf{S}_k\mathbf{S}_k^{\top}\mathbf{B}_k^{\top} \otimes \mathbf{E}^{\top}\mathbf{R}_k^{\top}\mathbf{\Lambda}_k^{-1}\mathbf{R}_k\mathbf{E})\right]^{-1}$$
$$\times \sum_{k=1}^{K}(\mathbf{B}_k\mathbf{S}_k \otimes \mathbf{E}^{\top}\mathbf{R}_k^{\top}\mathbf{\Lambda}_k^{-1})\mathbf{y}_k$$

and the Cramer-Rao lower bound for the estimator $\hat{\mathbf{a}}$, which is a lower bound on the covariance of $\hat{\mathbf{a}}$, is the inverse of $\mathcal{F}$. The FIM $\mathcal{F}$ is guaranteed to be invertible when, for at least one image, the matrix $\mathbf{B}_k\mathbf{S}_k\mathbf{S}_k^{\top}\mathbf{B}_k^{\top} \otimes \mathbf{E}^{\top}\mathbf{R}_k^{\top}\mathbf{\Lambda}_k^{-1}\mathbf{R}_k\mathbf{E}$ is full-rank.

The matrix $\mathbf{S}_k\mathbf{S}_k^{\top}$ has a rank of $N_k$ hence for $D_k > 1$ is rank-deficient. The blurring matrix $\mathbf{B}_k$ does not change the rank of the matrix that it multiplies from the right. In addition, as $\mathbf{\Lambda}_k^{-1}$ is full-rank, $\mathbf{E}^{\top}\mathbf{R}_k^{\top}\mathbf{\Lambda}_k^{-1}\mathbf{R}_k\mathbf{E}$ has a full rank of $M$ when the rows of $\mathbf{R}_k\mathbf{E}$ are at least as many as its columns, i.e., $L_k \geq M$. Therefore, $\mathbf{A}$ and consequently $\mathbf{X}$ is guaranteed to be uniquely identifiable given $\mathbf{Y}_k$, $k = 1, \dots, K$, only when at least one observed image, say the $q$th image, has full spatial resolution, i.e., $N_q = N$, with the number of its spectral bands being equal to or larger than the number of endmembers, i.e., $L_q \geq M$, so that, at least for the $q$th image, $\mathbf{B}_q\mathbf{S}_q\mathbf{S}_q^{\top}\mathbf{B}_q^{\top} \otimes \mathbf{E}^{\top}\mathbf{R}_q^{\top}\mathbf{\Lambda}_q^{-1}\mathbf{R}_q\mathbf{E}$ is full-rank.

In practice, it is rarely possible to satisfy the abovementioned requirements as multiband images with high spectral resolution are generally spatially downsampled and the number of bands of the ones with full spatial resolution, such as panchromatic or multispectral images, is often less than the number of endmembers. Hence, the inverse problem of recovering $\mathbf{A}$ from $\mathbf{Y}_k$, $k = 1, \dots, K$, is usually ill-posed or ill-conditioned. Thus, some prior knowledge must be injected into the estimation process to produce a unique and reliable estimate. The prior knowledge is intended to partially compensate for the information lost in spectral and spatial downsampling and usually stems from experimental evidence or common facts that may induce certain analytical properties or constraints. The prior information is commonly incorporated into the problem in the form of imposed constraints or additive regularization terms. Examples of prior knowledge about $\mathbf{A}$ that are regularly used in the literature are nonnegativity and sum-to-one constraints, matrix normal distribution with known or estimated parameters [36], sparse representation with a learned or known dictionary or basis [32], and minimal total variation [29].

### B. Regularization

To develop an algorithm for effective fusion of multiple multiband images with arbitrary spatial and spectral resolutions, we employ two mechanisms to regularize the maximum-likelihood cost function in (9).

As the first regularization mechanism, we impose a constraint on $\mathbf{A}$ such that its entries are nonnegative and sum to one in all columns. We express this constraint as $\mathbf{A} \geq 0$ and $\mathbf{1}_M^{\top}\mathbf{A} = \mathbf{1}_N^{\top}$ where $\mathbf{A} \geq 0$ means all the entries of $\mathbf{A}$ are greater than or equal to zero. As the second regularization mechanism, we add an isotropic vector total-variation penalty term, denoted by $\|\nabla\mathbf{A}\|_{2,1}$, to the cost function. Here, $\|\cdot\|_{2,1}$ is the $\ell_{2,1}$-norm operator that returns the sum of $\ell_2$-norms of all the columns of its matrix argument. In addition, we define

$$\nabla\mathbf{A} = \begin{bmatrix} \mathbf{A}\mathbf{D}_h \\ \mathbf{A}\mathbf{D}_v \end{bmatrix} \in \mathbb{R}^{2M \times N}$$

where $\mathbf{D}_h$ and $\mathbf{D}_v$ are discrete differential matrix operators that, respectively, yield the horizontal and vertical first-order backward differences (gradients) of the row-vectorized image that they multiply from the right. Consequently, we formulate our regularized optimization problem for estimating $\mathbf{A}$ as

$$\min_{\mathbf{A}} \frac{1}{2}\sum_{k=1}^{K}\left\|\mathbf{\Lambda}_k^{-1/2}(\mathbf{Y}_k - \mathbf{R}_k\mathbf{E}\mathbf{A}\mathbf{B}_k\mathbf{S}_k)\right\|_{\text{F}}^2 + \alpha\|\nabla\mathbf{A}\|_{2,1}$$
$$\text{subject to: } \mathbf{A} \geq 0 \text{ and } \mathbf{1}_M^{\top}\mathbf{A} = \mathbf{1}_N^{\top} \tag{10}$$

where $\alpha \geq 0$ is the regularization parameter.

The nonnegativity and sum-to-one constraints on $\mathbf{A}$, which force the columns of $\mathbf{A}$ to reside on the unit $(M-1)$-simplex, are naturally expected and help find a solution that is physically plausible. In addition, they implicitly induce sparseness in the solution. The total-variation penalty promotes solutions with a sparse gradient, a property that is known to be possessed by images of most natural scenes as they are usually made of piecewise homogeneous regions with few sudden changes at object boundaries or edges. Note that the subspace spanned by the endmembers is the one that the target image $\mathbf{X}$ lives in. Therefore, through the total-variation regularization of the abundance matrix $\mathbf{A}$, we regularize $\mathbf{X}$ indirectly.

## IV. Algorithm

Defining the set of values for $\mathbf{A}$ that satisfy the nonnegativity and sum-to-one constraints as

$$\mathcal{S} = \{\mathbf{A} | \mathbf{A} \geq 0, \mathbf{1}_M^\top \mathbf{A} = \mathbf{1}_N^\top\} \quad (11)$$

and making use of the indicator function $\iota_\mathcal{S}(\mathbf{A})$ defined as

$$\iota_\mathcal{S}(\mathbf{A}) = \begin{cases} 0 & \mathbf{A} \in \mathcal{S} \\ +\infty & \mathbf{A} \notin \mathcal{S}, \end{cases}$$

we rewrite (10) as

$$\min_{\mathbf{A}} \frac{1}{2} \sum_{k=1}^{K} \left\| \Lambda_k^{-1/2} (\mathbf{Y}_k - \mathbf{R}_k \mathbf{E} \mathbf{A} \mathbf{B}_k \mathbf{S}_k) \right\|_F^2 + \alpha \|\nabla \mathbf{A}\|_{2,1} + \iota_\mathcal{S}(\mathbf{A}). \quad (12)$$

### A. Iterations

We use the alternating direction method of multipliers (ADMM), also known as the split-Bregman method, to solve the convex but non-smooth optimization problem of (12). We split the problem to smaller and more manageable pieces by defining the auxiliary variables, $\mathbf{U}_k \in \mathbb{R}^{M \times N}$, $k = 1, ..., K$, $\mathbf{V} \in \mathbb{R}^{2M \times N}$, and $\mathbf{W} \in \mathbb{R}^{M \times N}$, and changing (12) into

$$\min_{\mathbf{A}, \mathbf{U}_1, ..., \mathbf{U}_K, \mathbf{V}, \mathbf{W}} \frac{1}{2} \sum_{k=1}^{K} \left\| \Lambda_k^{-1/2} (\mathbf{Y}_k - \mathbf{R}_k \mathbf{E} \mathbf{U}_k \mathbf{S}_k) \right\|_F^2 + \alpha \|\mathbf{V}\|_{2,1} + \iota_\mathcal{S}(\mathbf{W})$$

$$\text{subject to: } \mathbf{U}_k = \mathbf{A}\mathbf{B}_k, \mathbf{V} = \nabla\mathbf{A}, \mathbf{W} = \mathbf{A}. \quad (13)$$

Then, we write the augmented Lagrangian function associated with (13) as

$$\mathcal{L}(\mathbf{A}, \mathbf{U}_1, ..., \mathbf{U}_K, \mathbf{V}, \mathbf{W}, \mathbf{F}_1, ..., \mathbf{F}_K, \mathbf{G}, \mathbf{H})$$
$$= \frac{1}{2} \sum_{k=1}^{K} \left\| \Lambda_k^{-1/2} (\mathbf{Y}_k - \mathbf{R}_k \mathbf{E} \mathbf{U}_k \mathbf{S}_k) \right\|_F^2 + \alpha \|\mathbf{V}\|_{2,1} + \iota_\mathcal{S}(\mathbf{W})$$
$$+ \frac{\mu}{2} \sum_{k=1}^{K} \|\mathbf{A}\mathbf{B}_k - \mathbf{U}_k - \mathbf{F}_k\|_F^2 + \frac{\mu}{2} \|\nabla\mathbf{A} - \mathbf{V} - \mathbf{G}\|_F^2$$
$$+ \frac{\mu}{2} \|\mathbf{A} - \mathbf{W} - \mathbf{H}\|_F^2 \quad (14)$$

where $\mathbf{F}_k \in \mathbb{R}^{M \times N}$, $k = 1, ..., K$, $\mathbf{G} \in \mathbb{R}^{2M \times N}$, and $\mathbf{H} \in \mathbb{R}^{M \times N}$ are the scaled Lagrange multipliers and $\mu \geq 0$ is the penalty parameter.

Using the ADMM, we minimize the augmented Lagrangian function in an iterative fashion. At each iteration, we alternate the minimization with respect to the main unknown variable $\mathbf{A}$ and the auxiliary variables; then, we update the scaled Lagrange multipliers. Hence, we compute the iterates as

$$\mathbf{A}^{(n)} = \underset{\mathbf{A}}{\arg\min} \ \mathcal{L}\big(\mathbf{A}, \mathbf{U}_1^{(n-1)}, ..., \mathbf{U}_K^{(n-1)}, \mathbf{V}^{(n-1)}, \mathbf{W}^{(n-1)}, \mathbf{F}_1^{(n-1)}, ..., \mathbf{F}_K^{(n-1)}, \mathbf{G}^{(n-1)}, \mathbf{H}^{(n-1)}\big) \quad (15)$$

$$\left\{\mathbf{U}_1^{(n)}, ..., \mathbf{U}_K^{(n)}, \mathbf{V}^{(n)}, \mathbf{W}^{(n)}\right\}$$
$$= \underset{\mathbf{U}_1, ..., \mathbf{U}_K, \mathbf{V}, \mathbf{W}}{\arg\min} \ \mathcal{L}\big(\mathbf{A}^{(n)}, \mathbf{U}_1, ..., \mathbf{U}_K, \mathbf{V}, \mathbf{W}, \mathbf{F}_1^{(n-1)}, ..., \mathbf{F}_K^{(n-1)}, \mathbf{G}^{(n-1)}, \mathbf{H}^{(n-1)}\big) \quad (16)$$

$$\mathbf{F}_k^{(n)} = \mathbf{F}_k^{(n-1)} - (\mathbf{A}^{(n)}\mathbf{B}_k - \mathbf{U}_k^{(n)}), \ k = 1, ..., K$$
$$\mathbf{G}^{(n)} = \mathbf{G}^{(n-1)} - (\nabla\mathbf{A}^{(n)} - \mathbf{V}^{(n)})$$
$$\mathbf{H}^{(n)} = \mathbf{H}^{(n-1)} - (\mathbf{A}^{(n)} - \mathbf{W}^{(n)})$$

where superscript $(n)$ denotes the value of an iterate at iteration number $n \geq 0$. We repeat the iterations until convergence is reached up to a maximum allowed number of iterations.

Since we define the auxiliary variables independent of each other, the minimization of the augmented Lagrangian function (14) with respect to the auxiliary variables can be realized separately. Thus, (16) is equivalent to

$$\mathbf{U}_k^{(n)} = \underset{\mathbf{U}_k}{\arg\min} \ \frac{1}{2} \left\| \Lambda_k^{-1/2} (\mathbf{Y}_k - \mathbf{R}_k \mathbf{E} \mathbf{U}_k \mathbf{S}_k) \right\|_F^2 + \frac{\mu}{2} \left\| \mathbf{A}^{(n)}\mathbf{B}_k - \mathbf{U}_k - \mathbf{F}_k^{(n-1)} \right\|_F^2, k = 1, ..., K \quad (17)$$

$$\mathbf{V}^{(n)} = \underset{\mathbf{V}}{\arg\min} \ \alpha\|\mathbf{V}\|_{2,1} + \frac{\mu}{2} \left\| \nabla\mathbf{A}^{(n)} - \mathbf{V} - \mathbf{G}^{(n-1)} \right\|_F^2 \quad (18)$$

$$\mathbf{W}^{(n)} = \underset{\mathbf{W}}{\arg\min} \ \iota_\mathcal{S}(\mathbf{W}) + \frac{\mu}{2} \left\| \mathbf{A}^{(n)} - \mathbf{W} - \mathbf{H}^{(n-1)} \right\|_F^2. \quad (19)$$

### B. Solutions of subproblems

Considering (14), (15) can be written as

$$\mathbf{A}^{(n)} = \underset{\mathbf{A}}{\arg\min} \ \sum_{k=1}^{K} \left\| \mathbf{A}\mathbf{B}_k - \mathbf{U}_k^{(n-1)} - \mathbf{F}_k^{(n-1)} \right\|_F^2$$
$$+ \left\| \nabla\mathbf{A} - \mathbf{V}^{(n-1)} - \mathbf{G}^{(n-1)} \right\|_F^2$$
$$+ \left\| \mathbf{A} - \mathbf{W}^{(n-1)} - \mathbf{H}^{(n-1)} \right\|_F^2. \quad (20)$$

Calculating the gradient of the cost function in (20) with respect to $\mathbf{A}$ and setting it to zero gives

$$\mathbf{A}^{(n)} = \left[ \sum_{k=1}^{K} (\mathbf{U}_k^{(n-1)} + \mathbf{F}_k^{(n-1)})\mathbf{B}_k^\top + \mathbf{Q}_1^{(n-1)}\mathbf{D}_h^\top \right.$$
$$\left. + \mathbf{Q}_2^{(n-1)}\mathbf{D}_v^\top + \mathbf{W}^{(n-1)} + \mathbf{H}^{(n-1)} \right]$$
$$\times \left( \sum_{k=1}^{K} \mathbf{B}_k\mathbf{B}_k^\top + \mathbf{D}_h\mathbf{D}_h^\top + \mathbf{D}_v\mathbf{D}_v^\top + \mathbf{I}_N \right)^{-1} \quad (21)$$

where, for the convenience of presentation, we define $\mathbf{Q}_1^{(n-1)}$ and $\mathbf{Q}_2^{(n-1)}$ as

$$\begin{bmatrix} \mathbf{Q}_1^{(n-1)} \\ \mathbf{Q}_2^{(n-1)} \end{bmatrix} = \mathbf{V}^{(n-1)} + \mathbf{G}^{(n-1)}.$$

To make the computation of $\mathbf{A}^{(n)}$ in (21) more efficient, we assume that the two-dimensional convolutions represented by $\mathbf{B}_k$, $k = 1, \ldots, K$, are cyclic. In addition, we assume that the differential matrix operators $\mathbf{D}_h$ and $\mathbf{D}_v$ apply with periodic boundaries. Consequently, multiplications by $\mathbf{B}_k^{\top}$, $\mathbf{D}_h^{\top}$, and $\mathbf{D}_v^{\top}$ as well as by $(\sum_{k=1}^{K} \mathbf{B}_k \mathbf{B}_k^{\top} + \mathbf{D}_h \mathbf{D}_h^{\top} + \mathbf{D}_v \mathbf{D}_v^{\top} + \mathbf{I}_N)^{-1}$ can be performed through the use of the fast Fourier transform (FFT) algorithm and the circular convolution theorem. This theorem states that the Fourier transform of a circular convolution is the pointwise product of the Fourier transforms, i.e., a circular convolution can be expressed as the inverse Fourier transform of the product of the individual spectra [55].

Equating the gradient of the cost function in (17) with respect to $\mathbf{U}_k$ to zero results in

$$\mathbf{E}^{\top} \mathbf{R}_k^{\top} \mathbf{\Lambda}_k^{-1} \mathbf{R}_k \mathbf{E} \mathbf{U}_k^{(n)} \mathbf{S}_k \mathbf{S}_k^{\top} + \mu \mathbf{U}_k^{(n)}$$
$$= \mathbf{E}^{\top} \mathbf{R}_k^{\top} \mathbf{\Lambda}_k^{-1} \mathbf{Y}_k \mathbf{S}_k^{\top} + \mu \left( \mathbf{A}^{(n)} \mathbf{B}_k - \mathbf{F}_k^{(n-1)} \right). \quad (22)$$

Multiplying both sides of (22) from the right by the masking matrix $\mathbf{M}_k = \mathbf{S}_k \mathbf{S}_k^{\top}$ and its complement $\mathbf{I}_N - \mathbf{M}_k$ yields

$$\mathbf{U}_k^{(n)} \mathbf{M}_k = (\mathbf{E}^{\top} \mathbf{R}_k^{\top} \mathbf{\Lambda}_k^{-1} \mathbf{R}_k \mathbf{E} + \mu \mathbf{I}_N)^{-1}$$
$$\times \left[ \mathbf{E}^{\top} \mathbf{R}_k^{\top} \mathbf{\Lambda}_k^{-1} \mathbf{Y}_k \mathbf{S}_k^{\top} + \mu \left( \mathbf{A}^{(n)} \mathbf{B}_k - \mathbf{F}_k^{(n-1)} \right) \mathbf{M}_k \right] \quad (23)$$

and

$$\mathbf{U}_k^{(n)} (\mathbf{I}_N - \mathbf{M}_k) = \left( \mathbf{A}^{(n)} \mathbf{B}_k - \mathbf{F}_k^{(n-1)} \right) (\mathbf{I}_N - \mathbf{M}_k), \quad (24)$$

respectively. Note that we have $\mathbf{S}_k^{\top} \mathbf{S}_k = \mathbf{I}_N$ and $\mathbf{M}_k$ is idempotent, i.e., $\mathbf{M}_k \mathbf{M}_k = \mathbf{M}_k$. Summing both sides of (23) and (24) gives the solution of (17) for $k = 1, \ldots, K$ as

$$\mathbf{U}_k^{(n)} = \mathbf{U}_k^{(n)} \mathbf{M}_k + \mathbf{U}_k^{(n)} (\mathbf{I}_N - \mathbf{M}_k)$$
$$= (\mathbf{E}^{\top} \mathbf{R}_k^{\top} \mathbf{\Lambda}_k^{-1} \mathbf{R}_k \mathbf{E} + \mu \mathbf{I}_N)^{-1}$$
$$\times \left[ \mathbf{E}^{\top} \mathbf{R}_k^{\top} \mathbf{\Lambda}_k^{-1} \mathbf{Y}_k \mathbf{S}_k^{\top} + \mu \left( \mathbf{A}^{(n)} \mathbf{B}_k - \mathbf{F}_k^{(n-1)} \right) \mathbf{M}_k \right]$$
$$+ \left( \mathbf{A}^{(n)} \mathbf{B}_k - \mathbf{F}_k^{(n-1)} \right) (\mathbf{I}_N - \mathbf{M}_k).$$

The terms $(\mathbf{E}^{\top} \mathbf{R}_k^{\top} \mathbf{\Lambda}_k^{-1} \mathbf{R}_k \mathbf{E} + \mu \mathbf{I}_N)^{-1}$ and $\mathbf{E}^{\top} \mathbf{R}_k^{\top} \mathbf{\Lambda}_k^{-1} \mathbf{Y}_k \mathbf{S}_k^{\top}$ do not change during the iterations and can be precomputed.

The subproblem (18) can be decomposed pixelwise and its solution is linked to the so-called Moreau proximity operator of the $\ell_{2,1}$-norm given by column-wise vector-soft-thresholding [56], [57]. If we define

$$\mathbf{Z}^{(n)} = \nabla \mathbf{A}^{(n)} - \mathbf{G}^{(n-1)},$$

the $j$th column of $\mathbf{V}^{(n)}$, denoted by $\mathbf{v}_j^{(n)}$, is given in terms of the $j$th column of $\mathbf{Z}^{(n)}$, denoted by $\mathbf{z}_j^{(n)}$, as

$$\mathbf{v}_j^{(n)} = \frac{\max \left\{ \left\| \mathbf{z}_j^{(n)} \right\|_2 - \frac{\alpha}{\mu}, 0 \right\}}{\left\| \mathbf{z}_j^{(n)} \right\|_2} \mathbf{z}_j^{(n)}.$$

The solution of (19) is the value of the proximity operator of the indicator function $\iota_{\mathcal{S}}(\mathbf{W})$ at the point $\mathbf{A}^{(n)} - \mathbf{H}^{(n-1)}$, which is the projection of $\mathbf{A}^{(n)} - \mathbf{H}^{(n-1)}$ onto the set $\mathcal{S}$ defined by (11). Therefore, we have

$$\mathbf{W}^{(n)} = \underset{\mathbf{W} \in \mathcal{S}}{\operatorname{argmin}} \left\| \mathbf{A}^{(n)} - \mathbf{H}^{(n-1)} - \mathbf{W} \right\|_{\mathrm{F}}^2$$
$$= \Pi_{\mathcal{S}} \left\{ \mathbf{A}^{(n)} - \mathbf{H}^{(n-1)} \right\}$$

where $\Pi_{\mathcal{S}}\{\cdot\}$ denotes the projection onto $\mathcal{S}$. We implement this projection onto the unit $(M - 1)$-simplex employing the algorithm proposed in [58].

We present a summary of the proposed algorithm in Table I.

### C. Convergence

By defining

$$\mathcal{U} = [\mathbf{U}_1, \cdots, \mathbf{U}_K, \mathbf{V}, \mathbf{W}]^{\top}$$

and

$$\mathcal{C} = [\mathbf{B}_1, \cdots, \mathbf{B}_K, \mathbf{D}_h, \mathbf{D}_v, \mathbf{I}_N]^{\top},$$

(13) can be expressed as

$$\begin{aligned} \min_{\mathbf{A}} \quad & f(\mathcal{U}) \\ \text{subject to} \quad & \mathcal{U} = \mathcal{C} \mathbf{A}^{\top} \end{aligned} \quad (25)$$

where

$$f(\mathcal{U}) = \frac{1}{2} \sum_{k=1}^{K} \left\| \mathbf{\Lambda}_k^{-1/2} (\mathbf{Y}_k - \mathbf{R}_k \mathbf{E} \mathbf{U}_k \mathbf{S}_k) \right\|_{\mathrm{F}}^2 + \alpha \|\mathbf{V}\|_{2,1} + \iota_{\mathcal{S}}(\mathbf{W}).$$

The function $f(\mathcal{U})$ is closed, proper, and convex as it is a sum of closed, proper, and convex functions and $\mathcal{C}$ has full column rank. Therefore, according to [47, Theorem 8], if (25) has a solution, the proposed algorithm converges to this solution, regardless of the initial values as long as the penalty parameter $\mu$ is positive. If no solution exists, at least one of $\mathbf{A}^{(n)}$ and $\mathcal{U}^{(n)}$ will diverge.

## V. SIMULATIONS

To examine the performance of the proposed algorithm in comparison with the state-of-the-art, we simulate the fusion of three multiband images, viz. a panchromatic image, a multispectral image, and a hyperspectral image. To this end, we adopt the popular practice known as the Wald's protocol [78], which is to use a reference image with high spatial and spectral resolutions to generate the lower-resolution images that are fused and evaluate the fusion performance by comparing the fused image with the reference image.

We obtain the reference images of our experiments by cropping five publicly available hyperspectral images to the spatial resolutions given in Table II. These images are called Botswana[4], Indian Pines [59], Washington DC Mall[5], Moffett Field[6], and Kennedy Space Center[4]. The Botswana image has been captured by the Hyperion sensor aboard the Earth Observing 1 (EO-1) satellite, the Washington DC Mall image by the airborne-mounted Hyperspectral Digital Imagery Collection Experiment (HYDICE), and the Indian Pines, Moffett Filed, and Kennedy Space Center images by the NASA Airborne Visible/Infrared Imaging Spectrometer (AVIRIS) instrument. All images cover the visible near-infrared (VNIR) and short-wavelength infrared (SWIR) ranges with uncalibrated, noisy, and water-absorption bands removed. The spectral resolution of each image is also given in Table II. The data as well as the MATLAB code used to produce the results of this paper can be found at [60].

---

[4] http://www.ehu.eus/ccwintco/?title=Hyperspectral_Remote_Sensing_Scenes
[5] https://engineering.purdue.edu/~biehl/MultiSpec/hyperspectral.html

[6] http://aviris.jpl.nasa.gov/data/free_data.html

We generate three multiband images (panchromatic, multispectral, and hyperspectral) using each reference image. We obtain the hyperspectral images by applying a rotationally-symmetric 2D Gaussian blur filter with a kernel size of $13 \times 13$ and a standard deviation of 2.12 to each reference image followed by downsampling with a ratio of 4 in both spatial dimensions for all bands. For the multispectral images, we use a Gaussian blur filter with a kernel size of $7 \times 7$ and a standard deviation of 1.06 and downsampling with a ratio of 2 in both spatial dimensions for all bands of each reference image. Afterwards, we downgrade the resultant images spectrally by applying the spectral responses of the Landsat 8 multispectral sensor. This sensor has eight multispectral bands and one panchromatic band. Fig. 1 depicts the spectral responses of all the bands of this sensor[7]. We create the panchromatic images from the reference images using the panchromatic band of the Landsat 8 sensor without applying any spatial blurring or downsampling. We add zero-mean Gaussian white noise to each band of the produced multiband images such that the band-specific signal-to-noise ratio (SNR) is 30 dB for the multispectral and hyperspectral images and 40 dB for the panchromatic image. Note that we have selected the standard deviations of the abovementioned 2D Gaussian blur filters such that the normalized magnitude of the modulation transfer function (MTF) of both filters is approximately 0.25 at the Nyquist frequency in both spatial dimensions [77] as shown in Fig. 2.

The current multiband image fusion algorithms published in the literature are designed to fuse two images at a time. In order to compare the performance of the proposed algorithm with the state-of-the-art, we consider fusing the abovementioned three multiband images in three different ways, which we refer to as Pan + HS, Pan + (MS + HS), and (Pan + MS) + HS, using the existing algorithms for pansharpening, hyperspectral pansharpening, and hyperspectral-multispectral fusion. In Pan + HS, we only fuse the panchromatic and hyperspectral images. In Pan + (MS + HS), and (Pan + MS) + HS, we fuse the given images in two cascading stages. In Pan + (MS + HS), first, we fuse the multispectral and hyperspectral images. Then, we fuse the resultant hyperspectral image with the panchromatic image. We use the same algorithm at both stages, albeit with different parameter values. In (Pan + MS) + HS, we first fuse the panchromatic image with the multispectral one. Then, we fuse the pansharpened multispectral image with the hyperspectral image. We use two different algorithms at each of the two stages resulting in four combined solutions.

For pansharpening, which is the fusion of a panchromatic image with a multispectral one, we use two algorithms called the band-dependent spatial detail (BDSD) [61] and the modulation-transfer-function generalized Laplacian pyramid with high-pass modulation (MTF-GLP-HPM) [62]-[64]. The BDSD algorithm belongs to the class of component substitution methods and the MTF-GLP-HPM algorithm falls into the category of multiresolution analysis. In [12], where several pansharpening algorithms are studied, it is shown that the BDSD and MTF-GLP-HPM algorithms exhibit the best performance among all the considered ones.

For fusing a panchromatic or multispectral image with a hyperspectral image, we use two algorithms proposed in [29] and [65], [66], which are called HySure and R-FUSE-TV, respectively. These algorithms are based on total-variation regularization and are among the best performing and most efficient hyperspectral pansharpening and multispectral-hyperspectral fusion algorithms currently available [15], [67].

We use three performance metrics for assessing the quality of a fused image with respect to its reference image. The metrics are the relative dimensionless global error in synthesis (ERGAS)[8] [68], spectral angle mapper (SAM) [69], and $Q2^n$ [70]. The metric $Q2^n$ is a generalization of the universal image quality index (UIQI) proposed in [71] and an extension of the $Q4$ index [72] to hyperspectral images based on hypercomplex numbers.

We extract the endmembers (columns of $\mathbf{E}$) from each hyperspectral image using the vertex component analysis (VCA) algorithm [73]. The VCA is a fast unsupervised unmixing algorithm that assumes the endmembers as the vertices of a simplex encompassing the hyperspectral data cloud. We utilize the SUnSAL algorithm [74] together with the extracted endmembers to unmix each hyperspectral image and obtain its abundance matrix. Then, we upscale the resulting matrix by a factor of four and apply two-dimensional spline interpolation on each of its rows (abundance bands) to generate the initial estimate for the abundance matrix $\mathbf{A}^{(0)}$. We initialize the proposed algorithm as well as the HySure and R-FUSE-TV algorithms by this matrix.

To make our comparisons fair, we tune the values of the parameters in the HySure and R-FUSE-TV algorithms to yield the best possible performance in all experiments. In addition, in order to use the BDSD and MTF-GLP-HPM algorithms to their best potential, we provide these algorithms with the true point-spread function, i.e., the blurring kernel, used to generate the multispectral images.

Apart from the number of endmembers, which can be estimated using, for example, the HySime algorithm [23], the proposed algorithm has two tunable parameters, the total-variation regularization parameter $\alpha$ and the ADMM penalty parameter $\mu$. The automatic tuning of the values of these parameters is an interesting and challenging subject. There are a number of strategies that can be employed such as those proposed in [75] and [76]. We found through experimentations that although the value of $\mu$ impacts the convergence speed of the proposed algorithm, as long as it is within an appropriate range, it has little influence on the accuracy of the proposed algorithm. Therefore, we set it to $\mu = 1.5 \times 10^3$ in all experiments. The value of $\alpha$ affects the performance of the proposed algorithm in subtle ways as shown in Fig. 3 where we plot the performance metrics, ERGAS, SAM, and $Q2^n$, against $\alpha$ for the Botswana and Washington DC Mall images. The results in Fig. 3 suggest that, for different values of $\alpha$, there is a trade-off between the performance metrics, specifically,

---

[7] http://landsat.gsfc.nasa.gov/landsat-8/

[8] The original phrase in French is: erreur relative globale adimensionnelle de synthèse.

ERGAS and $Q2^n$ on one side and SAM on the other. Therefore, we tune the value of $\alpha$ for each experiment only roughly to obtain a reasonable set of values for all three performance metrics. We give the values of $\alpha$ used in the proposed algorithm in Table II.

In Table III, we give the values of the performance metrics to assess the quality of the images fused using the proposed algorithm and the considered benchmarks. We provide the performance metrics for the case of considering only the bands within the spectrum of the panchromatic image as well as the case of considering all bands, i.e., the entire spectrum of the reference image. We also give the time taken by each algorithm to produce the fused images[9]. According to the results in Table III, the proposed algorithm significantly outperforms the considered benchmarks. It is also evident from the required processing times that the computational (time) complexity of the proposed algorithm is lower than those of its contenders.

In Figs. 3 and 4, we plot the sorted per-pixel normalized root mean-square error (NRMSE) values of the proposed algorithm and the best performing algorithms from each of the Pan + HS, Pan + (MS + HS), and (Pan + MS) + HS categories. Fig. 4 corresponds to the case of considering only the spectrum of the panchromatic image and Fig. 5 to the case of considering the entire spectrum. We define the per-pixel NRMSE as

$$\frac{\left\| \mathbf{x}_j - \hat{\mathbf{x}}_j \right\|_2}{\left\| \mathbf{x}_j \right\|_2}$$

where $\mathbf{x}_j$ and $\hat{\mathbf{x}}_j$ are the $j$th column of the reference image $\mathbf{X}$ and the fused image $\hat{\mathbf{X}}$, respectively. We sort the NRMSE values in the ascending order.

In Fig. 5, we show RGB renderings of the reference images together with the panchromatic, multispectral, and hyperspectral images generated from them and used for the fusion. We also show the fused images yielded by the proposed algorithm and Pan + (MS + HS) fusion using the HySure algorithm, which generally performs better than the other considered benchmarks. The multispectral images are depicted using their red, green, and blue bands. The RGB representations of the hyperspectral images are rendered through transforming the spectral data to the CIE XYZ color space and then transforming the XYZ values to the sRGB color space. From visual inspection of the reference and fused images shown in Fig. 6, it is observed that the images fused by the proposed algorithm match their corresponding reference images better than the ones produced by the Pan + (MS + HS) fusion using the HySure algorithm do.

## VI. CONCLUSION

We proposed a new image fusion algorithm that can simultaneously fuse multiple multiband images. We utilized the well-known forward observation model together with the linear mixture model to cast the fusion problem as a reduced-dimension linear inverse problem. We used a vector total-variation penalty as well as nonnegativity and sum-to-one constraints on the endmember abundances to regularize the associated maximum-likelihood estimation problem. The regularization encourages the estimated fused image to have low rank with a sparse representation in the spectral domain while preserving the edges and discontinuities in the spatial domain. We solved the regularized problem using the alternating direction method of multipliers. We demonstrated the advantages of the proposed algorithm in comparison with the state-of-the-art via experiments with five real hyperspectral images that were done following the Wald's protocol.

## REFERENCES

[1] R. Arablouei, "Fusion of multiple multiband images with complementary spatial and spectral resolutions," *Int. Conf. Acoust., Speech Signal Process.*, Calgary, Canada, Apr. 2018.

[2] A. Mohammadzadeh, A. Tavakoli, and M. J. V. Zoej, "Road extraction based on fuzzy logic and mathematical morphology from pansharpened IKONOS images," *Photogramm. Rec.*, vol. 21, no. 113, pp. 44–60, Feb. 2006.

[3] C. Souza, Jr, L. Firestone, L. M. Silva, and D. Roberts, "Mapping forest degradation in the Eastern amazon from SPOT 4 through spectral mixture models," *Remote Sens. Environ.*, vol. 87, no. 4, pp. 494–506, 2003.

[4] G. A. Licciardi, A. Villa, M. M. Khan, and J. Chanussot, "Image fusion and spectral unmixing of hyperspectral images for spatial improvement of classification maps," in *Proc. IEEE Int. Conf. Geoscience Remote Sensing*, 2012, pp. 7290–729.

[5] J. M. Bioucas-Dias, A. Plaza, N. Dobigeon, M. Parente, Q. Du, P. Gader, and J. Chanussot, "Hyperspectral unmixing overview: Geometrical, statistical, and sparse regression-based approaches," *IEEE J. Select. Topics Appl. Earth Observ. Remote Sens.*, vol. 5, no. 2, pp. 354–379, Apr. 2012.

[6] J. M. Bioucas-Dias *et al.*, "Hyperspectral remote sensing data analysis and future challenges," *IEEE Geosci. Remote Sens. Mag.*, vol. 1, no. 2, pp. 6–36, Jun. 2013.

[7] G. A. Shaw and H.-H. K. Burke, "Spectral imaging for remote sensing," *Lincoln Lab. J.*, vol. 14, no. 1, pp. 3–28, 2003.

[8] J. Greer, "Sparse demixing of hyperspectral images," *IEEE Trans. Image Process.*, vol. 21, no. 1, pp. 219–228, Jan. 2012.

[9] R. Arablouei and F. de Hoog, "Hyperspectral image recovery via hybrid regularization," *IEEE Trans. Image Process.*, vol. 25, no. 12, pp. 5649–5663, Dec. 2016.

[10] R. Arablouei, "Spectral unmixing with perturbed endmembers," *IEEE Trans. Geosci. Remote Sens.*, submitted for publication, 2018; available online at https://osf.io/wgkxd/.

[11] R. Arablouei, E. Goan, S. Gensemer, and B. Kusy, "Fast and robust push-broom hyperspectral imaging via DMD-based scanning," in *Proc. SPIE 9948, Novel Optical Systems Design and Optimization XIX*, San Diego, CA, USA, Aug./Sep. 2016, id. 99480A.

[12] G. Vivone *et al.*, "A critical comparison among pansharpening algorithms," *IEEE Trans. Geosci. Remote Sens.*, vol. 53, no. 5, pp. 2565–2586, May 2015.

[13] L. Alparone, L. Wald, J. Chanussot, C. Thomas, P. Gamba, and L. M. Bruce, "Comparison of pansharpening algorithms: Outcome of the 2006 GRS-S data-fusion contest," *IEEE Trans. Geosci. Remote Sens.*, vol. 45, no. 10, pp. 3012–3021, Oct. 2007.

[14] B. Aiazzi, L. Alparone, S. Baronti, A. Garzelli, and M. Selva, "25 years of pansharpening: A critical review and new developments," in *Signal and Image Processing for Remote Sensing*, C. H. Chen, Ed., 2nd ed. Boca Raton, FL, USA: CRC Press, 2011, ch. 28, pp. 533–548.

[15] L. Loncan *et al.*, "Hyperspectral pansharpening: A review," *IEEE Geosci. Remote Sens. Mag.*, vol. 3, no. 3, pp. 27–46, Sep. 2015.

[16] M. Selva, B. Aiazzi, F. Butera, L. Chiarantini, and S. Baronti, "Hyper-sharpening: A first approach on SIM-GA data," *IEEE J. Select. Topics Appl. Earth Observ. Remote Sens.*, vol. 8, no. 6, pp. 3008–3024, Jun. 2015.

[17] M. T. Eismann and R. C. Hardie, "Application of the stochastic mixing model to hyperspectral resolution enhancement," *IEEE Trans. Geosci. Remote Sens.*, vol. 42, no. 9, pp. 1924–1933, Sep. 2004.

---

[9] We used MATLAB with a 2.9GHz Core-i7 CPU and 24GB of DDR3 RAM and ran each of the proposed, HySure, and R-FUSE-TV algorithms for 200 iterations as they always converged sufficiently after this number of iterations.

[18] B. Huang, H. Song, H. Cui, J. Peng, and Z. Xu, "Spatial and spectral image fusion using sparse matrix factorization," *IEEE Trans. Geosci. Remote Sens.*, vol. 52, no. 3, pp. 1693–1704, Mar. 2014.

[19] Y. Zhang, S. De Backer, and P. Scheunders, "Noise-resistant wavelet-based Bayesian fusion of multispectral and hyperspectral images," *IEEE Trans. Geosci. Remote Sens.*, vol. 47, no. 11, pp. 3834–3843, Nov. 2009.

[20] R. Kawakami, J. Wright, Y.-W. Tai, Y. Matsushita, M. Ben-Ezra, and K. Ikeuchi, "High-resolution hyperspectral imaging via matrix factorization," in *Proc. IEEE Conf. Comput. Vis. Pattern Recognit. (CVPR)*, Jun. 2011, pp. 2329–2336.

[21] E. Wycoff, T.-H. Chan, K. Jia, W.-K. Ma, and Y. Ma, "A non-negative sparse promoting algorithm for high resolution hyperspectral imaging," in Proc. *IEEE Int. Conf. Acoustics, Speech Signal Process.*, Vancouver, Canada, May 2013, pp. 1409–1413.

[22] K. Cawse-Nicholson, S. Damelin, A. Robin, and M. Sears, "Determining the intrinsic dimension of a hyperspectral image using random matrix theory," *IEEE Trans. Image Process.*, vol. 22, no. 4, pp. 1301–1310, Apr. 2013.

[23] J. Bioucas-Dias and J. Nascimento, "Hyperspectral subspace identification," *IEEE Trans. Geosci. Remote Sens.*, vol. 46, no. 8, pp. 2435–2445, Aug. 2008.

[24] D. Landgrebe, "Hyperspectral image data analysis," *IEEE Signal Process. Mag.*, vol. 19, no. 1, pp. 17–28, Jan. 2002.

[25] R. Zurita-Milla, J. G. P. W. Clevers, and M. E. Schaepman, "Unmixing-based landsat TM and MERIS FR data fusion," *IEEE Geosci. Remote Sens. Lett.*, vol. 5, no. 3, pp. 453–457, Jul. 2008.

[26] L. Rudin, S. Osher, and E. Fatemi, "Nonlinear total variation based noise removal algorithms," *Physica D: Nonlinear Phenomena*, vol. 60, pp. 259–268, Nov. 1992.

[27] A. Beck and M. Teboulle, "Fast gradient-based algorithms for constrained total variation image denoising and deblurring problems," *IEEE Trans. Image Process.*, vol. 18, no. 11, pp. 2419–2434, Nov. 2009.

[28] X. Bresson and T. Chan, "Fast dual minimization of the vectorial total variation norm and applications to color image processing," *Inverse Problems and Imaging*, vol. 2, no. 4, pp. 455–484, Nov. 2008.

[29] M. Simões, J. Bioucas-Dias, L. B. Almeida, and J. Chanussot, "A convex formulation for hyperspectral image superresolution via subspace-based regularization," *IEEE Trans. Geosci. Remote Sens.*, vol. 53, no. 6, pp. 3373–3388, Jun. 2015.

[30] X. He, L. Condat, J. Bioucas-Dias, J. Chanussot, and J. Xia, "A new pansharpening method based on spatial and spectral sparsity priors," *IEEE Trans. Image Process.*, vol. 23, no. 9, pp. 4160–4174, Sep. 2014.

[31] F. Palsson, J. R. Sveinsson, and M. O. Ulfarsson, "A new pansharpening algorithm based on total variation," *IEEE Geosci. Remote Sens. Lett.*, vol. 11, no. 1, pp. 318-322, Jan. 2014.

[32] Q. Wei, J. Bioucas-Dias, N. Dobigeon, and J. Tourneret, "Hyperspectral and multispectral image fusion based on a sparse representation," *IEEE Trans. Geosci. Remote Sens.*, vol. 53, no. 7, pp. 3658–3668, Jul. 2015.

[33] W. Dong *et al.*, "Hyperspectral image super-resolution via non-negative structured sparse representation," *IEEE Trans. Image Process.*, vol. 25, no. 5, pp. 2337–2352, May 2016.

[34] C. Grohnfeldt, X. X. Zhu, and R. Bamler, "Jointly sparse fusion of hyperspectral and multispectral imagery," in *Proc. IEEE Int. Geosci. Remote Sens. Symp.*, Melbourne, Australia, Jul. 2013, pp. 4090–4093.

[35] N. Akhtar, F. Shafait, and A. Mian, "Bayesian sparse representation for hyperspectral image super resolution," in *Proc. IEEE Conf. Comput. Vision Pattern Recognition*, Boston, MA, USA, Jun. 2015, pp. 3631–3640.

[36] Q. Wei, N. Dobigeon, and J.-Y. Tourneret, "Bayesian fusion of multiband images," *IEEE J. Sel. Topics Signal Process.*, vol. 9, no. 6, pp. 1117–1127, Sep. 2015.

[37] R. C. Hardie, M. T. Eismann, and G. L. Wilson, "MAP estimation for hyperspectral image resolution enhancement using an auxiliary sensor," *IEEE Trans. Image Process.*, vol. 13, no. 9, pp. 1174–1184, Sep. 2004.

[38] Y. Zhang, A. Duijster, and P. Scheunders, "A Bayesian restoration approach for hyperspectral images," *IEEE Trans. Geosci. Remote Sens.*, vol. 50, no. 9, pp. 3453–3462, Sep. 2012.

[39] M. A. Veganzones, M. Simões, G. Licciardi, N. Yokoya, J. M. Bioucas-Dias, and J. Chanussot, "Hyperspectral super-resolution of locally low rank images from complementary multisource data," *IEEE Trans. Image Process.*, vol. 25, no. 1, pp. 274–288, Jan. 2016.

[40] O. Berne, A. Helens, P. Pilleri, and C. Joblin, "Non-negative matrix factorization pansharpening of hyperspectral data: An application to mid-infrared astronomy," in *Proc. IEEE GRSS Workshop Hyperspectral Image Signal Process.: Evolution in Remote Sens.*, Reykjavik, Iceland, Jun. 2010, pp. 1–4.

[41] N. Yokoya, T. Yairi, and A. Iwasaki, "Coupled nonnegative matrix factorization unmixing for hyperspectral and multispectral data fusion," *IEEE Trans. Geosci. Remote Sens.*, vol. 50, no. 2, pp. 528–537, Feb. 2012.

[42] C. Lanaras, E. Baltsavias, and K. Schindler, "Hyperspectral superresolution by coupled spectral unmixing," in *Proc. IEEE ICCV*, Santiago, Chile, Dec. 2015, pp. 3586–3594.

[43] Q. Wei, J. Bioucas-Dias, N. Dobigeon, J.-Y. Tourneret, M. Chen, and S. Godsill, "Multiband image fusion based on spectral unmixing," *IEEE Trans. Geosci. Remote Sens.*, vol. 54, no. 12, pp. 7236- 7249, Dec. 2016.

[44] C. Kwan, B. Budavari, A. C. Bovik, and G. Marchisio, "Blind quality assessment of fused WorldView-3 images by using the combinations of pansharpening and hypersharpening paradigms," *IEEE Geosci. Remote Sens. Lett.*, vol. 14, no. 10, pp. 1835–1839, Oct. 2017.

[45] N. Yokoya, T. Yairi, and A. Iwasaki, "Hyperspectral, multispectral, and panchromatic data fusion based on coupled non-negative matrix factorization," in *Proc. IEEE GRSS Workshop Hyperspectral Image Signal Process.: Evolution in Remote Sens.*, Lisbon, Portugal, Jun. 2011, pp. 1–4.

[46] M. Afonso, J. M. Bioucas-Dias, and M. Figueiredo, "An augmented Lagrangian approach to the constrained optimization formulation of imaging inverse problems," *IEEE Trans. Image Process.*, vol. 20, no. 3, pp. 681–95, Mar. 2011.

[47] J. Eckstein and D. P. Bertsekas, "On the Douglas–Rachford splitting method and the proximal point algorithm for maximal monotone operators," *Math. Program.*, vol. 55, nos. 1–3, pp. 293–318, 1992.

[48] D. Gabay and B. Mercier, "A dual algorithm for the solution of nonlinear variational problems via finite-element approximation," *Comput. Math. Appl.*, vol. 2, no. 1, pp. 17–40, 1976.

[49] R. Glowinski and A. Marroco, "Sur l'approximation, par éléments finis d'ordre un, et la résolution, par pénalisation-dualité d'une classe de problèmes de Dirichlet non linéaires," *Revue française d'automatique, informatique, recherche opérationnelle. Analyse numérique*, vol. 9, no. 2, pp. 41–76, 1975.

[50] S. Boyd, N. Parikh, E. Chu, B. Peleato, and J. Eckstein, "Distributed optimization and statistical learning via the alternating direction method of multipliers," *Foundation and Trends in Machine Learning*, vol. 3, no. 1, pp. 1–122, 2011.

[51] E. Esser, "Applications of Lagrangian-based alternating direction methods and connections to split-Bregman," Center Comput. Appl. Math., Univ. California, Los Angeles, Tech. Rep. 09-31, 2009.

[52] B.-C. Gao, M. J. Montes, C. O. Davis, and A. F. Goetz, "Atmospheric correction algorithms for hyperspectral remote sensing data of land and ocean," *Remote Sensing Environ.*, vol. 113, pp. S17–S24, Sept. 2009.

[53] A. Zare and K. C. Ho, "Endmember variability in hyperspectral analysis," *IEEE Signal Process. Mag.*, vol. 31, no. 1, pp. 95–104, Jan. 2014.

[54] N. Dobigeon, J.-Y. Tourneret, C. Richard, J. C. M. Bermudez, S. McLaughlin, and A. O. Hero, "Nonlinear unmixing of hyperspectral images: Models and algorithms," *IEEE Signal Process. Mag.*, vol. 31, no. 1, pp. 89–94, Jan. 2014.

[55] T. G. Stockham, Jr., "High-speed convolution and correlation," in *Proc. ACM Spring Joint Computer Conf.*, New York, NY, USA, Apr. 1966, pp. 229-233.

[56] D. Donoho and I. Johnstone, "Adapting to unknown smoothness via wavelet shrinkage," *Journal of the American Statistical Association*, vol. 90, no. 432, pp. 1200–1224, Dec. 1995.

[57] P. Combettes and J.-C. Pesquet, "Proximal splitting methods in signal processing," in *Fixed-Point Algorithms for Inverse Problems in Science and Engineering*, New York, NY, USA: Springer-Verlag, 2011, pp. 185–212.

[58] L. Condat, "Fast projection onto the simplex and the $\ell_1$ ball," *Mathematical Programming, Series A*, vol. 158, pp. 575–585, 2016.

[59] M. Baumgardner, L. Biehl, and D. Landgrebe, "220 band AVIRIS hyperspectral image data set: June 12, 1992 Indian pine test site 3," Purdue University Research Repository, 2015. doi:10.4231/R7RX991C

[60] https://github.com/Reza219/Multiple-multiband-image-fusion

[61] A. Garzelli, F. Nencini, and L. Capobianco, "Optimal MMSE pan sharpening of very high resolution multispectral images," *IEEE Trans. Geosci. Remote Sens.*, vol. 46, no. 1, pp. 228–236, Jan. 2008.

[62] B. Aiazzi, L. Alparone, S. Baronti, A. Garzelli, and M. Selva, "MTF-tailored multiscale fusion of high-resolution MS and Pan imagery," *Photogrammetric Engineering and Remote Sensing*, vol. 72, no. 5, pp. 591–596, May 2006.

[63] J. Lee and C. Lee, "Fast and efficient panchromatic sharpening," *IEEE Trans. Geosci. Remote Sens.*, vol. 48, no. 1, pp. 155–163, Jan. 2010.

[64] B. Aiazzi, L. Alparone, S. Baronti, A. Garzelli, and M. Selva, "An MTF-based spectral distortion minimizing model for pan-sharpening of very high resolution multispectral images of urban areas," in *Proc. 2nd GRSS/ISPRS Joint Workshop Remote Sens. Data Fusion URBAN Areas*, Berlin, Germany, May 2003, pp. 90–94.

[65] Q. Wei, N. Dobigeon, J.-Y. Tourneret, J. M. Bioucas-Dias, and S. Godsill, "R-FUSE: Robust fast fusion of multiband images based on solving a Sylvester equation," *IEEE Signal Process. Lett.*, vol. 23, no. 11, pp. 1632-1636, Nov. 2016.

[66] Q. Wei, N. Dobigeon, and J.-Y. Tourneret, "Fast fusion of multi-band images based on solving a Sylvester equation," *IEEE Trans. Image Process.*, vol. 24, no. 11, pp. 4109–4121, Nov. 2015.

[67] N. Yokoya, C. Grohnfeldt, and J. Chanussot, "Hyperspectral and multispectral data fusion: A comparative review of the recent literature," *IEEE Geosci. Remote Sens. Mag.*, vol. 5, no. 2, pp. 29–56, Jun. 2017.

[68] L. Wald, "Quality of high resolution synthesised images: Is there a simple criterion?" in *Proc. Int. Conf. Fusion Earth Data*, Nice, France, Jan. 2000, pp. 99–103.

[69] F. A. Kruse *et al.*, "The spectral image processing system (SIPS): Interactive visualization and analysis of imaging spectrometer data," *Remote Sens. Environ.*, vol. 44, no. 2–3, pp. 145–163, May–Jun 1993.

[70] A. Garzelli and F. Nencini, "Hypercomplex quality assessment of multi/hyperspectral images," *IEEE Geosci. Remote Sens. Lett.*, vol. 6, no. 4, pp. 662–665, Oct. 2009.

[71] Z. Wang and A. C. Bovik, "A universal image quality index," *IEEE Signal Process. Lett.*, vol. 9, no. 3, pp. 81–84, Mar. 2002.

[72] L. Alparone, S. Baronti, A. Garzelli, and F. Nencini, "A global quality measurement of pan-sharpened multispectral imagery," *IEEE Geosci. Remote Sens. Lett.*, vol. 1, no. 4, pp. 313–317, Oct. 2004.

[73] J. Nascimento and J. Bioucas-Dias, "Vertex component analysis: A fast algorithm to unmix hyperspectral data," *IEEE Trans. Geosci. Remote Sens.*, vol. 43, no. 4, pp. 898–910, Apr. 2005.

[74] J. Bioucas-Dias and M. Figueiredo, "Alternating direction algorithms for constrained sparse regression: Application to hyperspectral unmixing," in *Proc. IEEE GRSS Workshop Hyperspectral Image Signal Process.: Evolution in Remote Sens.*, Reykjavik, Iceland, Jun. 2010, vol. 1, pp. 1–4.

[75] D. Donoho and I. Johnstone, "Adapting to unknown smoothness via wavelet shrinkage," *J. Amer. Stat. Assoc.*, vol. 90, no. 432, pp. 1200–1224, Dec. 1995.

[76] G. Golub, M. Heath, and G. Wahba, "Generalized cross-validation as a method for choosing a good ridge parameter," *Technometrics*, vol. 21, no. 2, pp. 215–223, May 1979.

[77] L. Alparone, S. Baronti, B. Aiazzi, and A. Garzelli, "Spatial methods for multispectral pansharpening: Multiresolution analysis demystified," *IEEE Trans. Geosci. Remote Sens.*, vol. 54, no. 5, pp. 2563–2576, May 2016.

[78] L. Wald, T. Ranchin, and M. Mangolini, "Fusion of satellite images of different spatial resolutions: Assessing the quality of resulting image," *IEEE Trans. Geosci. Remote Sens.*, vol. 43, pp. 1391–1402, 2005.

Table I
THE PROPOSED ALGORITHM

---

initialize

$\mathbf{E} \leftarrow \text{VCA}(\mathbf{Y}_l)$ % if $\mathbf{E}$ is not known and $\mathbf{Y}_l$ has full spectral resolution

$\mathbf{A}^{(0)} \leftarrow$ upscale and interpolate the output of $\text{SUnSAL}(\mathbf{Y}_l, \mathbf{E})$

for $k = 1, \ldots, K$

$\quad \mathbf{U}_k^{(0)} = \mathbf{A}^{(0)}$

$\quad \mathbf{F}_k^{(n)} = \mathbf{0}_{M \times N}$

$\mathbf{V}^{(0)} = \mathbf{A}^{(0)}, \mathbf{W}^{(0)} = \mathbf{A}^{(0)}$

$\mathbf{G}^{(0)} = \mathbf{0}_{M \times N}, \mathbf{H}^{(0)} = \mathbf{0}_{M \times N}$

for $n = 1, 2, \ldots$ % until a convergence criterion is met or a given maximum number of iterations is reached

$$\begin{bmatrix} \mathbf{Q}_1^{(n-1)} \\ \mathbf{Q}_2^{(n-1)} \end{bmatrix} = \mathbf{V}^{(n-1)} + \mathbf{G}^{(n-1)}$$

$$\mathbf{A}^{(n)} = \left[ \sum_{k=1}^{K} (\mathbf{U}_k^{(n-1)} + \mathbf{F}_k^{(n-1)}) \mathbf{B}_k^\top + \mathbf{Q}_1^{(n-1)} \mathbf{D}_h^\top + \mathbf{Q}_2^{(n-1)} \mathbf{D}_v^\top + \mathbf{W}^{(n-1)} + \mathbf{H}^{(n-1)} \right] \left( \sum_{k=1}^{K} \mathbf{B}_k \mathbf{B}_k^\top + \mathbf{D}_h \mathbf{D}_h^\top + \mathbf{D}_v \mathbf{D}_v^\top + \mathbf{I}_N \right)^{-1}$$

for $k = 1, \ldots, K$

$\quad \mathbf{U}_k^{(n)} = (\mathbf{E}^\top \mathbf{R}_k^\top \boldsymbol{\Lambda}_k^{-1} \mathbf{R}_k \mathbf{E} + \mu \mathbf{I}_N)^{-1} [\mathbf{E}^\top \mathbf{R}_k^\top \boldsymbol{\Lambda}_k^{-1} \mathbf{Y}_k \mathbf{S}_k^\top + \mu (\mathbf{A}^{(n)} \mathbf{B}_k - \mathbf{F}_k^{(n-1)}) \mathbf{M}_k] + (\mathbf{A}^{(n)} \mathbf{B}_k - \mathbf{F}_k^{(n-1)})(\mathbf{I}_N - \mathbf{M}_k)$

$\mathbf{Z}^{(n)} = \nabla \mathbf{A}^{(n)} - \mathbf{G}^{(n-1)}$

for $j = 1, \ldots, N$

$$\quad \mathbf{v}_j^{(n)} = \frac{\max \left\{ \left\| \mathbf{z}_j^{(n)} \right\|_2 - \frac{\alpha}{\mu}, 0 \right\}}{\left\| \mathbf{z}_j^{(n)} \right\|_2} \mathbf{z}_j^{(n)}$$

$\mathbf{W}^{(n)} = \Pi_S \{ \mathbf{A}^{(n)} - \mathbf{H}^{(n-1)} \}$

for $k = 1, \ldots, K$

$\quad \mathbf{F}_k^{(n)} = \mathbf{F}_k^{(n-1)} - (\mathbf{A}^{(n)} \mathbf{B}_k - \mathbf{U}_k^{(n)})$

$\mathbf{G}^{(n)} = \mathbf{G}^{(n-1)} - (\nabla \mathbf{A}^{(n)} - \mathbf{V}^{(n)})$

$\mathbf{H}^{(n)} = \mathbf{H}^{(n-1)} - (\mathbf{A}^{(n)} - \mathbf{W}^{(n)})$

calculate the fused image

$\hat{\mathbf{X}} = \mathbf{E} \mathbf{A}^{(n)}$

---

Table II
THE SPATIAL AND SPECTRAL DIMENSIONS OF THE CONSIDERED REFERENCE HYPERSPECTRAL IMAGES AND THE VALUE OF THE REGULARIZATION PARAMETER USED IN THE PROPOSED ALGORITHM WITH EACH IMAGE

| image | no. of rows | no. of columns | no. of bands | $\alpha$ |
|---|---|---|---|---|
| Botswana | 400 | 240 | 145 | 5 |
| Indian Pines | 400 | 400 | 200 | 7 |
| Washington DC Mall | 400 | 300 | 191 | 5 |
| Moffett Field | 480 | 320 | 176 | 22 |
| Kennedy Space Center | 500 | 400 | 176 | 28 |

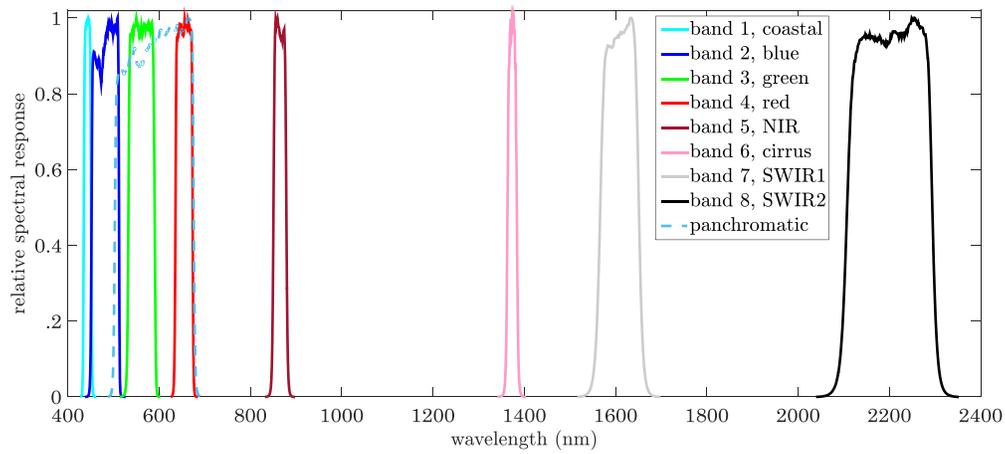

Fig. 1. The spectral responses of the Landsat 8 multispectral and panchromatic sensors.

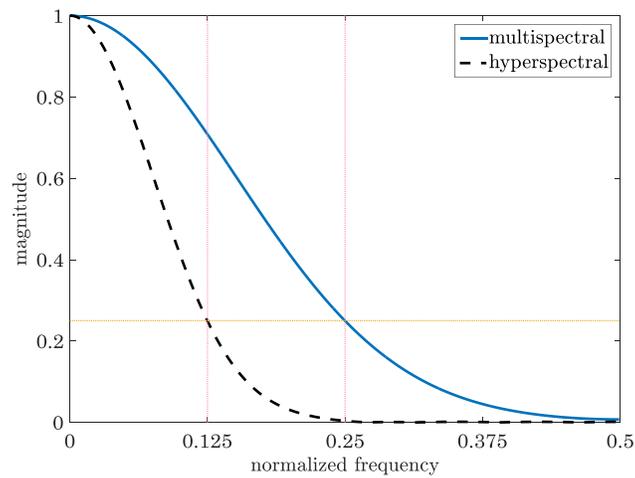

Fig. 2. The modulation transfer function (normalized spatial-frequency response) of the used 2D Gaussian blur filters in both spatial dimensions. The solid curve corresponds to the filter used to generate the multispectral images and the dashed curve corresponds to the filter used to generate the hyperspectral images.

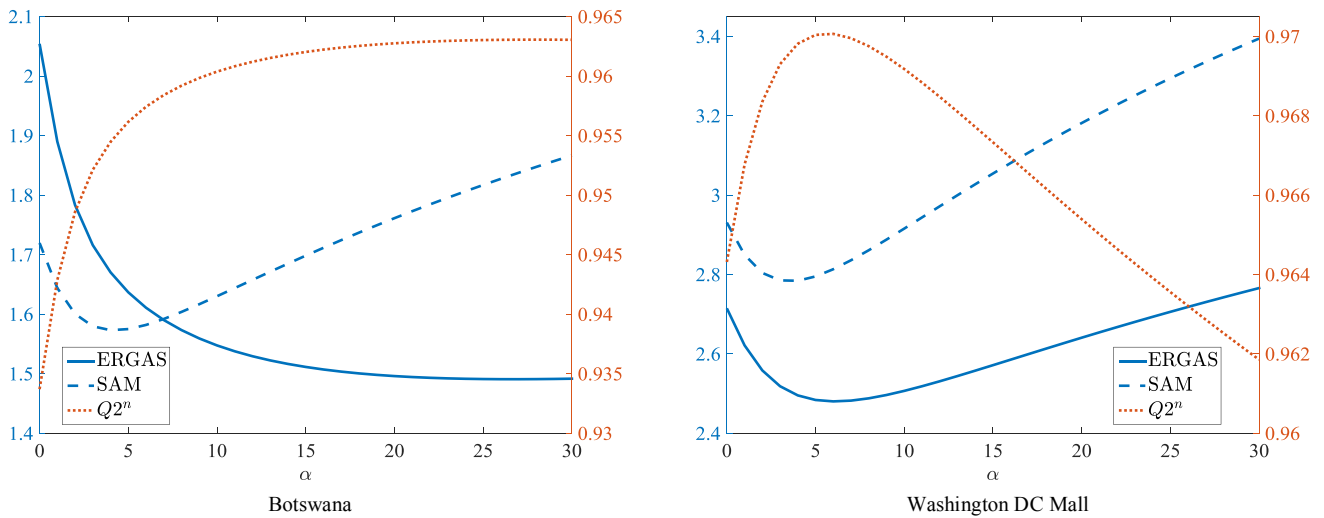

Fig. 3. The values of the performance metrics versus the regularization parameter $\alpha$ for the experiments with Botswana and Washington DC Mall images. The left $y$-axis corresponds to ERGAS and SAM and the right $y$-axis to $Q2^n$.

Table III
THE VALUES OF THE PERFORMANCE METRICS FOR ASSESSING THE FUSION QUALITY AS WELL AS THE RUNTIMES OF THE CONSIDERED ALGORITHMS FOR EXPERIMENTS WITH DIFFERENT IMAGES

Botswana

| fusion | algorithm(s) | spectrum of panchromatic image | | | entire spectrum | | | time (s) |
|---|---|---|---|---|---|---|---|---|
| | | ERGAS | SAM (°) | $Q2^n$ | ERGAS | SAM (°) | $Q2^n$ | |
| Pan + MS + HS | proposed | 0.900 | 1.355 | 0.980 | 1.637 | 1.575 | 0.956 | 47.01 |
| Pan + HS | HySure | 1.273 | 1.975 | 0.967 | 1.839 | 2.435 | 0.946 | 61.20 |
| | R-FUSE-TV | 1.272 | 1.974 | 0.967 | 1.840 | 2.436 | 0.946 | 61.17 |
| Pan + (MS + HS) | HySure | 1.256 | 1.721 | 0.962 | 1.992 | 2.101 | 0.937 | 78.28 |
| | R-FUSE-TV | 1.265 | 1.734 | 0.961 | 2.002 | 2.113 | 0.937 | 79.44 |
| (Pan + MS) + HS | BDSD & HySure | 1.393 | 1.971 | 0.955 | 2.458 | 2.359 | 0.912 | 62.58 |
| | BDSD & R-FUSE-TV | 1.392 | 1.971 | 0.956 | 2.461 | 2.365 | 0.912 | 62.10 |
| | MTF-GLP-HPM & HySure | 1.441 | 2.120 | 0.957 | 2.181 | 2.442 | 0.931 | 62.78 |
| | MTF-GLP-HPM & R-FUSE-TV | 1.440 | 2.124 | 0.957 | 2.185 | 2.446 | 0.931 | 62.20 |

Indian Pines

| fusion | algorithm(s) | spectrum of panchromatic image | | | entire spectrum | | | time (s) |
|---|---|---|---|---|---|---|---|---|
| | | ERGAS | SAM (°) | $Q2^n$ | ERGAS | SAM (°) | $Q2^n$ | |
| Pan + MS + HS | proposed | 0.304 | 0.293 | 0.990 | 0.500 | 0.761 | 0.969 | 80.21 |
| Pan + HS | HySure | 0.420 | 0.547 | 0.986 | 0.813 | 1.108 | 0.632 | 106.75 |
| | R-FUSE-TV | 0.425 | 0.555 | 0.986 | 0.813 | 1.113 | 0.632 | 106.47 |
| Pan + (MS + HS) | HySure | 0.656 | 0.641 | 0.961 | 0.834 | 1.117 | 0.594 | 134.79 |
| | R-FUSE-TV | 0.695 | 0.642 | 0.953 | 0.875 | 1.120 | 0.573 | 134.32 |
| (Pan + MS) + HS | BDSD & HySure | 0.538 | 0.517 | 0.972 | 0.803 | 1.183 | 0.670 | 108.33 |
| | BDSD & R-FUSE-TV | 0.539 | 0.520 | 0.972 | 0.794 | 1.182 | 0.674 | 107.34 |
| | MTF-GLP-HPM & HySure | 0.566 | 0.563 | 0.972 | 0.959 | 1.268 | 0.626 | 108.48 |
| | MTF-GLP-HPM & R-FUSE-TV | 0.567 | 0.567 | 0.972 | 0.947 | 1.270 | 0.628 | 107.51 |

Washington DC Mall

| fusion | algorithm(s) | spectrum of panchromatic image | | | entire spectrum | | | time (s) |
|---|---|---|---|---|---|---|---|---|
| | | ERGAS | SAM (°) | $Q2^n$ | ERGAS | SAM (°) | $Q2^n$ | |
| Pan + MS + HS | proposed | 0.731 | 1.116 | 0.997 | 2.484 | 2.795 | 0.970 | 59.52 |
| Pan + HS | HySure | 1.171 | 2.047 | 0.992 | 3.822 | 4.539 | 0.930 | 79.02 |
| | R-FUSE-TV | 1.171 | 2.042 | 0.992 | 3.832 | 4.537 | 0.930 | 78.38 |
| Pan + (MS + HS) | HySure | 0.937 | 1.718 | 0.994 | 3.233 | 3.592 | 0.949 | 99.74 |
| | R-FUSE-TV | 1.204 | 1.738 | 0.991 | 3.270 | 3.664 | 0.947 | 100.53 |
| (Pan + MS) + HS | BDSD & HySure | 1.114 | 2.039 | 0.992 | 4.174 | 5.048 | 0.918 | 79.68 |
| | BDSD & R-FUSE-TV | 1.104 | 2.060 | 0.992 | 4.251 | 5.033 | 0.916 | 78.41 |
| | MTF-GLP-HPM & HySure | 1.308 | 1.870 | 0.991 | 4.380 | 5.147 | 0.911 | 79.28 |
| | MTF-GLP-HPM & R-FUSE-TV | 1.298 | 1.884 | 0.991 | 4.440 | 5.114 | 0.910 | 78.13 |

Moffett Field

| fusion | algorithm(s) | spectrum of panchromatic image | | | entire spectrum | | | time (s) |
|---|---|---|---|---|---|---|---|---|
| | | ERGAS | SAM (°) | $Q2^n$ | ERGAS | SAM (°) | $Q2^n$ | |
| Pan + MS + HS | proposed | 0.572 | 0.786 | 0.992 | 4.232 | 3.148 | 0.885 | 77.37 |
| Pan + HS | HySure | 0.902 | 1.151 | 0.985 | 6.507 | 4.233 | 0.823 | 107.73 |
| | R-FUSE-TV | 0.914 | 1.152 | 0.984 | 6.416 | 4.210 | 0.827 | 106.20 |
| Pan + (MS + HS) | HySure | 0.826 | 1.004 | 0.986 | 5.078 | 3.603 | 0.868 | 134.78 |
| | R-FUSE-TV | 0.964 | 1.014 | 0.977 | 5.100 | 3.670 | 0.845 | 135.20 |
| (Pan + MS) + HS | BDSD & HySure | 1.061 | 1.135 | 0.980 | 5.325 | 4.065 | 0.829 | 108.91 |
| | BDSD & R-FUSE-TV | 1.058 | 1.134 | 0.980 | 5.244 | 4.039 | 0.834 | 106.12 |
| | MTF-GLP-HPM & HySure | 1.396 | 1.122 | 0.968 | 5.924 | 4.384 | 0.824 | 108.98 |
| | MTF-GLP-HPM & R-FUSE-TV | 1.396 | 1.123 | 0.969 | 5.835 | 4.360 | 0.830 | 106.28 |

Kennedy Space Center

| fusion | algorithm(s) | spectrum of panchromatic image | | | entire spectrum | | | time (s) |
|---|---|---|---|---|---|---|---|---|
| | | ERGAS | SAM (°) | $Q2^n$ | ERGAS | SAM (°) | $Q2^n$ | |
| Pan + MS + HS | proposed | 1.024 | 1.628 | 0.984 | 2.468 | 3.211 | 0.909 | 99.94 |
| Pan + HS | HySure | 1.451 | 2.426 | 0.979 | 3.544 | 3.995 | 0.890 | 138.16 |
| | R-FUSE-TV | 1.518 | 2.496 | 0.974 | 3.680 | 3.795 | 0.886 | 134.97 |
| Pan + (MS + HS) | HySure | 1.462 | 2.203 | 0.967 | 2.851 | 3.546 | 0.909 | 172.18 |
| | R-FUSE-TV | 1.875 | 2.343 | 0.939 | 2.986 | 4.155 | 0.878 | 172.25 |
| (Pan + MS) + HS | BDSD & HySure | 1.738 | 2.594 | 0.949 | 3.727 | 4.824 | 0.850 | 138.66 |
| | BDSD & R-FUSE-TV | 1.691 | 2.547 | 0.953 | 3.534 | 4.584 | 0.865 | 135.74 |
| | MTF-GLP-HPM & HySure | 6.801 | 3.250 | 0.912 | 9.532 | 5.183 | 0.805 | 138.60 |
| | MTF-GLP-HPM & R-FUSE-TV | 8.143 | 3.264 | 0.914 | 11.130 | 5.197 | 0.816 | 135.58 |

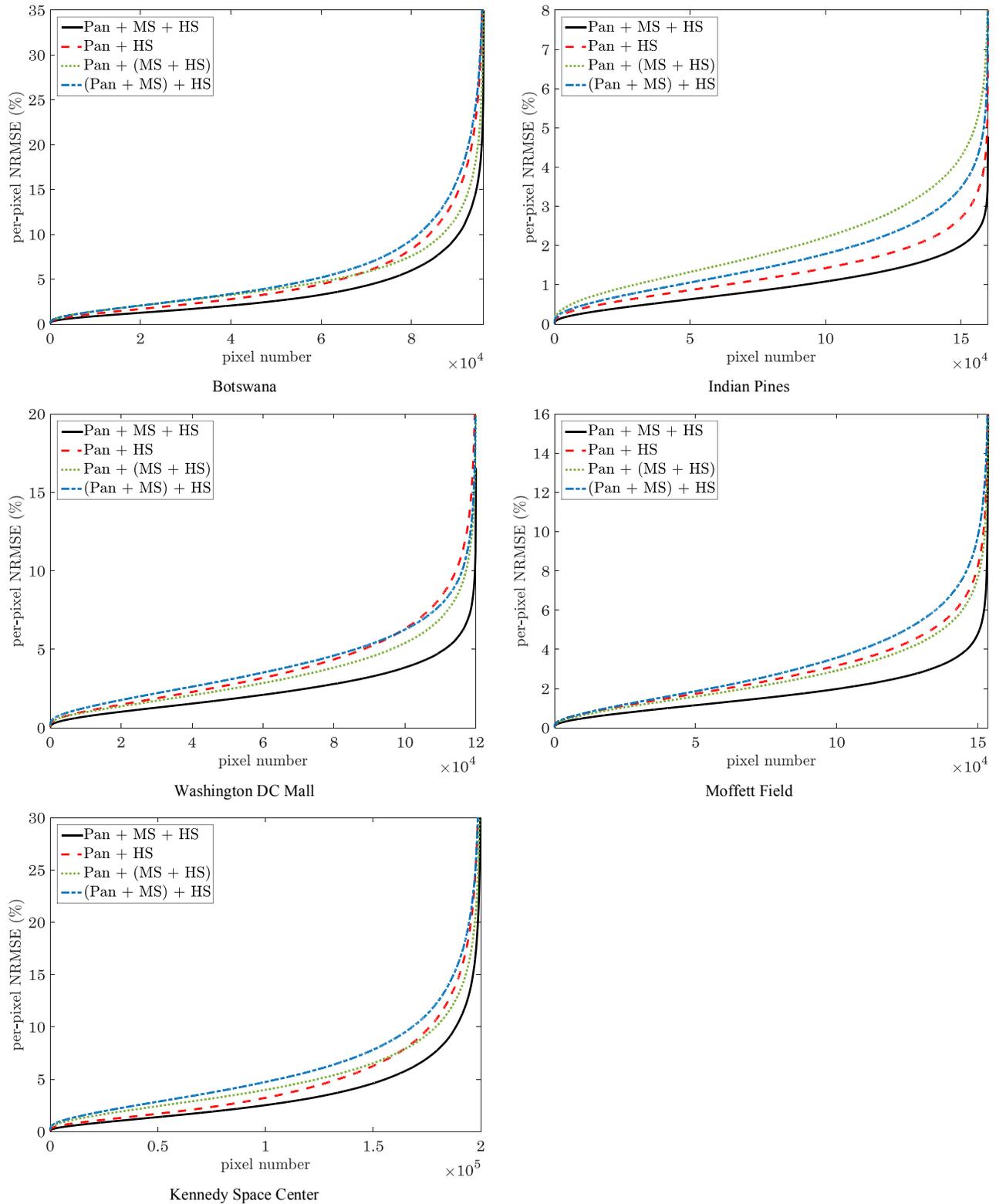

Fig. 4. The sorted per-pixel NRMSE of different algorithms measured only on the spectrum of the panchromatic image in experiments with different images.

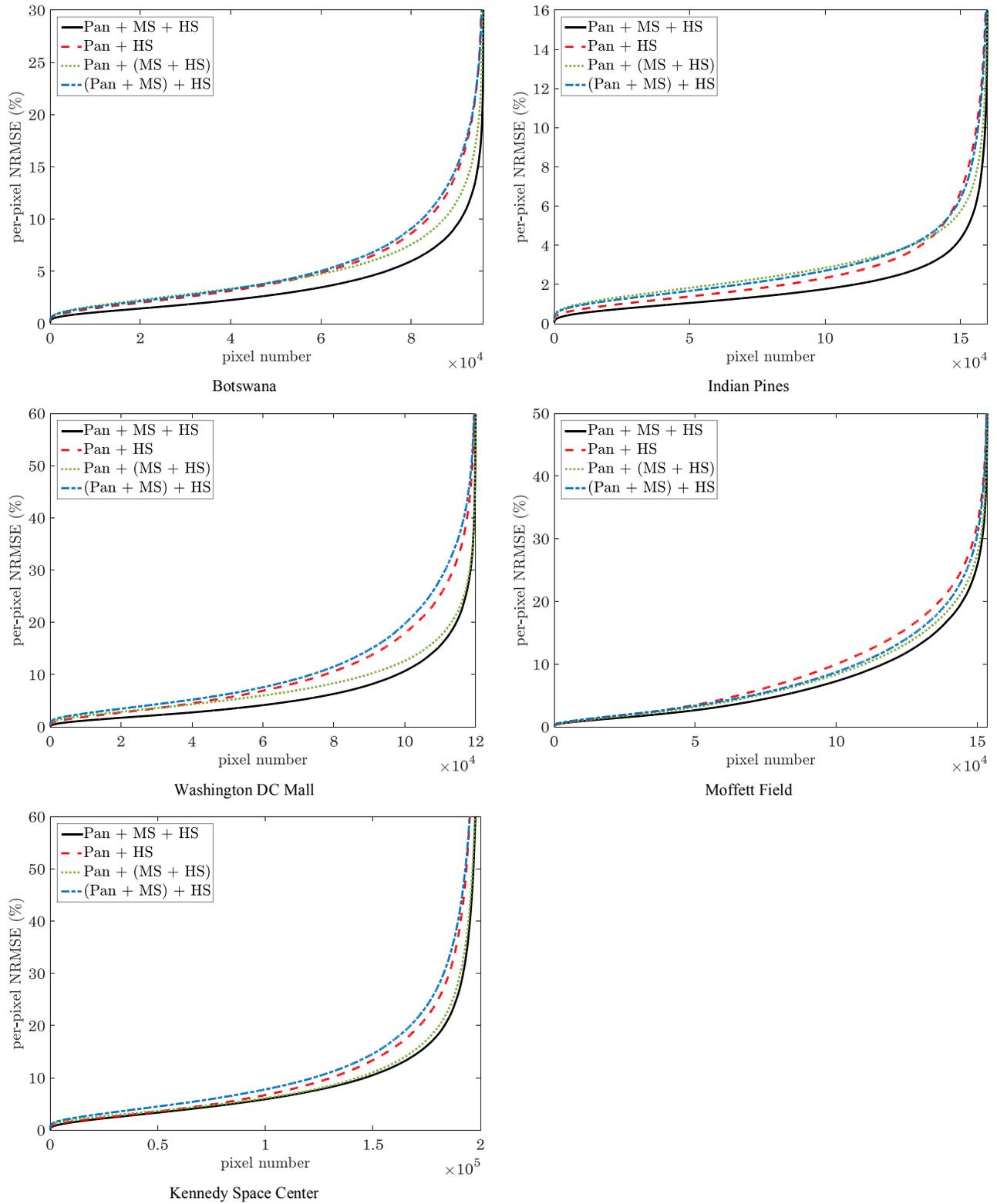

Fig. 5. The sorted per-pixel NRMSE of different algorithms measured on the entire spectrum in experiments with different images.

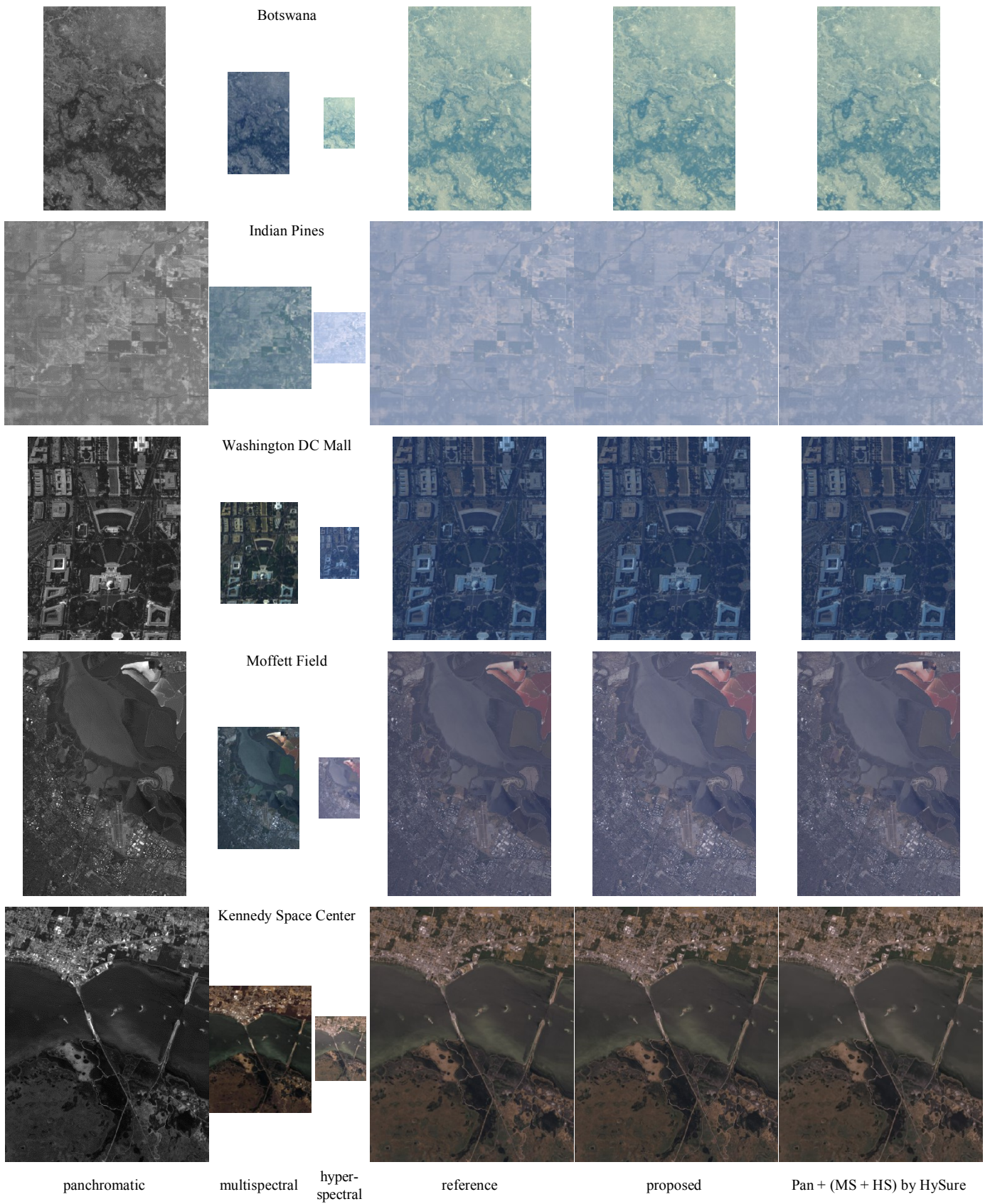

Botswana

Indian Pines

Washington DC Mall

Moffett Field

Kennedy Space Center

| panchromatic | multispectral | hyper-spectral | reference | proposed | Pan + (MS + HS) by HySure |

Fig. 6. The panchromatic, multispectral, and hyperspectral images that are fused together, the reference hyperspectral image, and the fused images produced by the proposed algorithm and the Pan + (MS + HS) method using the HySure algorithm.